\documentclass{article}
\usepackage{iclr2023_conference,times}

\usepackage[utf8]{inputenc}
\usepackage[T1]{fontenc}
\usepackage{url}
\usepackage{booktabs}
\usepackage{amsfonts}
\usepackage{nicefrac}
\usepackage{microtype}
\usepackage[hidelinks]{hyperref}
\hypersetup{
    colorlinks=true,
    linkcolor=blue,
    urlcolor=blue,
    citecolor=blue,
    anchorcolor=blue}
\usepackage{listings}
\usepackage[inline]{enumitem}
\usepackage[pdftex]{graphicx}
\usepackage{amsmath}
\usepackage{amssymb}
\usepackage{amsthm}
\usepackage{stmaryrd}
\usepackage{float}
\usepackage[export]{adjustbox}
\usepackage{caption}
\usepackage{subcaption}
\usepackage{mathtools, nccmath}
\usepackage{tikz}
\usepackage{multicol}
\usepackage{wrapfig}
\usepackage{algorithm}
\usepackage[algo2e, ruled,vlined, boxed, linesnumbered]{algorithm2e}
\SetArgSty{textnormal}
\usepackage{setspace}
\usepackage{amsmath,amsfonts,bm}
\usepackage{dsfont}

\def\1{\bm{1}}

\def\rve{{\mathbf{e}}}

\def\rvu{{\mathbf{i}}}

\def\rvu{{\mathbf{u}}}
\def\rvv{{\mathbf{v}}}

\def\rvx{{\mathbf{x}}}
\def\rvy{{\mathbf{y}}}
\def\rvz{{\mathbf{z}}}

\DeclareMathAlphabet{\mathsfit}{\encodingdefault}{\sfdefault}{m}{sl}
\SetMathAlphabet{\mathsfit}{bold}{\encodingdefault}{\sfdefault}{bx}{n}

\def\gD{{\mathcal{D}}}

\def\gM{{\mathcal{M}}}

\def\gX{{\mathcal{X}}}
\def\gY{{\mathcal{Y}}}

\newcommand{\R}{\mathbb{R}}

\DeclareMathOperator{\Proj}{Proj}

\newcommand{\defeq}{\vcentcolon=}

\def\bP{\mathbf{P}}

\def\ones{\mathds{1}}
\DeclarePairedDelimiterX{\dotp}[2]{\langle}{\rangle}{#1, #2}

\DeclareMathOperator*{\argmin}{argmin}

\newcommand\eqdef{\stackrel{\mathclap{\tiny\mbox{def.}}}{=}}

\newcommand{\dif}{\mathop{}\!\mathrm{d}}

\DeclareMathOperator*{\Exp}{\mathbb{E}}
\def\OT{\textup{OT}}
\def\OTe{\textup{OT}_{\varepsilon}}
\def\UBOT{\textup{UBOT}}

\def\diag{\textup{diag}}                      % Wasserstein Distance

\definecolor{darkblue}{HTML}{1A254B}
\definecolor{lightblue}{HTML}{A7BED3}
\definecolor{blue}{HTML}{2B50AA}
\definecolor{green}{HTML}{81B5AE}
\definecolor{pink}{HTML}{F2545B}
\definecolor{red}{HTML}{A4243B}
\definecolor{lightgray}{HTML}{E5EBEF}

\def\bm{\mathbf{m}}
\newcommand{\newinf}{\mathop{\mathrm{inf}\vphantom{\mathrm{sup}}}}
\lstdefinestyle{codestyle}{
    commentstyle=\color{blue},
    keywordstyle=\color{lightblue},
    numberstyle=\tiny\color{gray},
    stringstyle=\color{pink},
    basicstyle=\ttfamily\footnotesize,
    breakatwhitespace=false,
    breaklines=true,
    captionpos=b,
    keepspaces=true,
    showspaces=false,
    showstringspaces=false,
    showtabs=true,
    tabsize=2,
    frame=leftline
}
\title{Neural Unbalanced Optimal Transport \\ via Cycle-Consistent Semi-Couplings}

\author{Frederike L\"ubeck\thanks{These authors contributed equally to this work. Correspondence to: Frederike L\"ubeck (\texttt{fluebeck@ethz.ch}) and Charlotte Bunne (\texttt{bunnec@ethz.ch}).}$\;\,$\thanks{Work done while at Harvard University.} \\
ETH Zurich \hspace{58pt}  \\
\And
Charlotte Bunne\footnotemark[1] \\
ETH Zurich \\
\And
Gabriele Gut \\
University of Zurich \hspace{28pt} \\ 
\AND
Jacobo Sarabia del Castillo \\
University of Zurich \\
\And
Lucas Pelkmans \\
University of Zurich \\
\And
David Alvarez-Melis \\
Microsoft Research \\
}

\iclrfinalcopy 

\begin{document}

\maketitle

\begin{abstract}
\looseness -1 Comparing \emph{unpaired} samples of a distribution or population taken at different points in time is a fundamental task in many application domains where measuring populations is destructive and cannot be done repeatedly on the same sample, such as in single-cell biology. Optimal transport (OT) can solve this challenge by learning an optimal coupling of samples across distributions from unpaired data. However, the usual formulation of OT assumes conservation of mass, which is violated in \emph{unbalanced} scenarios in which the population size changes (e.g., cell proliferation or death) between measurements. In this work, we introduce \textsc{NubOT}, a \emph{neural unbalanced OT} formulation that relies on the formalism of \emph{semi-couplings} to account for creation and destruction of mass. To estimate such semi-couplings and generalize out-of-sample, we derive an efficient parameterization based on neural optimal transport maps and propose a novel algorithmic scheme through a cycle-consistent training procedure. We apply our method to the challenging task of forecasting heterogeneous responses of multiple cancer cell lines to various drugs, where we observe that by accurately modeling cell proliferation and death, our method yields notable improvements over previous neural optimal transport methods.
\end{abstract}

\section{Introduction}\label{sec:introduction}

Modeling change is at the core of various problems in the natural sciences, from  dynamical processes driven by natural forces to population trends induced by interventions. In all these cases, the gold standard is to track particles or individuals across time, which allows for immediate estimation of individual (or aggregate) effects. But maintaining these pairwise correspondences across interventions or time is not always possible, for example, when the same sample cannot be measured more than once. This is typical in biomedical sciences, where the process of measuring is often altering or destructive. For example, single-cell biology profiling methods destroy the cells and thus cannot be used repeatedly on the same cell. In these situations, one must rely on comparing \textit{different} replicas of a population and, absent a natural identification of elements across the populations, infer these correspondences from data in order to model evolution or intervention effects.

\looseness -1 The problem of inferring correspondences across unpaired samples in biology has been traditionally tackled by relying on average and aggregate perturbation responses \citep{green2016systems, zhan2019mek, sheldon2007collective} or by applying mechanistic or linear models  \citep{yuan2021cellbox, dixit2016perturb} in, potentially, a learned latent space \citep{lotfollahi2019scgen}. Cellular responses to treatments are, however, highly complex and heterogeneous. To effectively predict the drug response of a patient during treatment and capture such cellular heterogeneity, it is necessary to learn nonlinear maps describing such perturbation responses on the level of single cells.
Assuming perturbations incrementally alter molecular profiles of cells, such as gene expression or signaling activities, recent approaches have utilized optimal transport to predict changes and alignments \citep{schiebinger2019optimal, bunne2022recovering, tong2020trajectorynet}. By returning a coupling between control and perturbed cell states, which overall minimizes the cost of matching, optimal transport can solve that puzzle and reconstruct these incremental changes in cell states over time.
% stands out for its unique ability to simultaneously model correspondence and transformation.

Despite the advantages mentioned above, the classic formulation of OT is ill-suited to model processes where the population changes in \textit{size}, e.g., where elements might be created or destroyed over time. This is the case, for example, in single-cell biology, where interventions of interest typically promote proliferation of certain cells and death of others. Such scenarios violate the assumption of conservation of mass that the classic OT problem relies upon. Relaxing this assumption yields a generalized formulation, known as the \textit{unbalanced} OT (UBOT) problem, for which recent work has studied its properties \citep{liero2018optimal, chizat2018scaling}, numerical solution \citep{chapel2021unbalanced}, and has applied it successfully to problems in single cell biology \citep{yang2018scalable}. Yet, these methods typically scale poorly with sample size, are prone to unstable solutions, or make limiting assumptions, e.g., only allowing for destruction but not creation of mass. 

\begin{figure}[t]
    \centering
    \includegraphics[width=\textwidth]{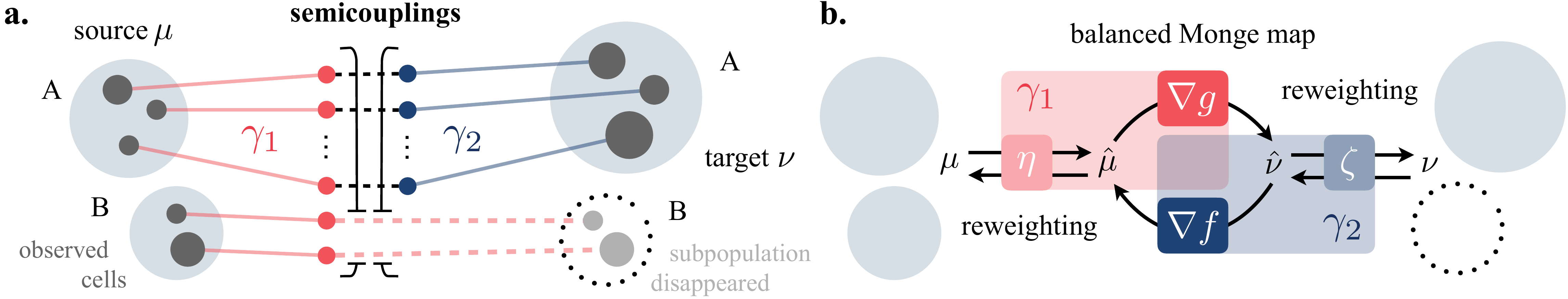}
    \caption{\looseness -1 \textbf{a.} A semi-coupling pair ({\color{pink} $\gamma_1$}, {\color{darkblue} $\gamma_2$}) consists of two couplings that together solve the \emph{unbalanced} OT problem. Intuitively, {\color{pink} $\gamma_1$} describes where mass goes as it leaves from $\mu$, and {\color{darkblue} $\gamma_2$} where it comes from as it arrives in $\nu$. \textbf{b.} \textsc{NubOT} parameterizes the semi-couplings ({\color{pink} $\gamma_1$}, {\color{darkblue} $\gamma_2$}) as the composition of reweighting functions $\eta$ and $\zeta$ and the dual potentials $f$ and $g$ between the then \emph{balanced} problem.}
    \label{fig:overview}
\end{figure}

In this work, we address these shortcomings by introducing a novel formulation of the unbalanced OT problem that relies on the formalism of semi-couplings introduced by \citet{chizat2018unbalanced}, while still obtaining an explicit transport map that models the transformation between distributions. The advantage of the latter is that it allows mapping new out-of-sample points, and it provides an interpretable characterization of the underlying change in distribution. Since the unbalanced OT problem does not directly admit a Monge (i.e., mapping-based) formulation, we propose to \textit{learn} to jointly `re-balance' the two distributions, thereby allowing us to estimate a map between their rescaled versions. To do so, we leverage prior work \citep{makkuva2020optimal, korotin2021neural} that learns the transport map as the gradient of a convex dual potential \citep{Brenier1987} parameterized via an input convex neural network \citep{amos2017input}. In addition, we derive a simple update rule to learn the rescaling functions. Put together, these components yield a reversible, parameterized, and computationally feasible implementation of the semi-coupling unbalanced OT formulation (Fig. ~\ref{fig:overview}). 

In short, the main \textbf{contributions} of this work are: \begin{enumerate*}[label=(\roman*)]
\item A novel formulation of the unbalanced optimal transport problem that weaves together the theoretical foundations of semi-couplings with the practical advantage of transport maps; 
\item A general, scalable, and efficient algorithmic implementation for this formulation based on dual potentials parameterized via convex neural network architectures; and
\item An empirical validation on the challenging task of predicting perturbation responses of single cells to multiple cancer drugs,  where our method successfully predicts cell proliferation and death, in addition to faithfully modeling the perturbation responses on the level of single cells.
\end{enumerate*}

\section{Background} \label{sec:background}

\subsection{Optimal Transport} \label{sec:background_OT}

\looseness -1 For two probability measures $\mu, \nu$ in $\mathcal{P}(\mathcal{X})$ with $\mathcal{X} = \mathbb{R}^d$ and a real-valued continuous cost function $c\in\mathcal{C}(\mathcal{X}^2)$, the optimal transport problem \citep{kantorovich1942transfer} is defined as
\begin{equation} \label{eq:ot}
    \OT(\mu, \nu) \defeq \inf_{\gamma\in \Gamma(\mu,\nu)}\int_{\gX^2} c(x,y) \gamma(dx, dy),
\end{equation}
where $\Gamma(\mu, \nu) = \{ \gamma \in \gM_+(\gX^2), \text{ s.t. } (\Proj_1)_\sharp \gamma = \mu, (\Proj_2)_\sharp \gamma = \nu  \}$ is the set of couplings in the cone of nonnegative Radon measures $\gM_+(\gX^2)$ with respective marginals $\mu, \nu$. When instantiated on finite discrete measures, such as $\mu=\sum_{i=1}^n u_i\delta_{\rvx_i}$ and $\nu=\sum_{j=1}^m v_j\delta_{\rvy_j}$, with $\mathbf{u} \in \Sigma_n, \mathbf{v} \in \Sigma_m$ this problem translates to a linear program, which can be regularized using an entropy term~\citep{Peyre2019computational}. For $\varepsilon\geq0$, set 
\begin{equation}\label{eq:reg-ot}
\OTe(\mu, \nu) \defeq \min_{\bP\in U(\rvu,\rvv)} \dotp{\bP}{[c(\rvx_i, \rvy_j)]_{ij}}  \,-\varepsilon H(\bP),
\end{equation}
where $H(\bP) \defeq -\sum_{ij} \bP_{ij} (\log \bP_{ij} - 1)$ and the polytope $U(\rvu, \rvv)$ is the set of matrices $\{\bP\in\mathbb{R}^{n \times m}_+, \bP\mathbf{1}_m =\rvu, \bP^\top\mathbf{1}_n=\rvv\}$. For clarity, we will sometimes write $\OTe(\rvu, \rvv, \{\rvx_i\}, \{\rvy_j\})$.
Notice that the definition above reduces to \eqref{eq:ot} when $\varepsilon=0$. Setting $\varepsilon>0$ yields a faster and differentiable proxy to approximate $\OT$ and allows fast numerical approximation via the Sinkhorn algorithm \citep{cuturi2013sinkhorn}, but introduces a bias, since in general $\OTe(\mu,\mu)\ne 0$.

\paragraph{Neural optimal transport.}
% \paragraph{Convexity}
To parameterize \eqref{eq:ot} and allow to predict how a measure evolves from $\mu$ to $\nu$, we introduce an alternative formulation known as the \citeauthor{Monge1781} problem (\citeyear{Monge1781}) given by
\begin{equation} \label{eq:monge}
    \OTe(\mu, \nu)=\inf _{T: T_{\sharp} \mu=\nu} \int_{\mathcal{X}}c(x, T(x)) d \mu(x),
\end{equation}
with pushforward operator $\sharp$ and the optimal solution $T^*$ known as the \citeauthor{Monge1781} map between $\mu$ and $\nu$.
\looseness -1 When cost $c$ is the quadratic Euclidean distance, i.e., $c = \| \cdot\|^2_2$, \citeauthor{Brenier1987}'s theorem (\citeyear{Brenier1987}) states that this \citeauthor{Monge1781} map is necessarily the gradient $\nabla \psi$ of a convex potential $\psi : \mathcal{X} \mapsto \mathbb{R}$ such that $\nabla \psi_\sharp \mu = \nu$, i.e., $T^*(x) = \nabla \psi(x)$.
This connection has far-reaching impact and is a central component of recent neural optimal transport solvers \citep{makkuva2020optimal, bunne2022proximal, alvarez2021optimizing, korotin2020wasserstein, bunne2022supervised, fan2021barycenter}.
Instead of (indirectly) learning the Monge map $T$ \citep{yang2018scalable, fan2021scalable}, it is sufficient to restrict the computational effort to learning a \emph{good} convex potential $\nabla_\theta$, parameterized via input convex neural networks (ICNN) \citep{amos2017input}, s.t. $\nabla_\theta \psi \sharp \mu = \nu$.
Alternatively, parameterizations of such maps can be carried out via the dual formulation of \eqref{eq:ot} \citep[Proposition 1.11, Theorem 1.39]{santambrogio2015optimal}, i.e.,
\begin{equation}
    \OT(\mu, \nu) \defeq \sup _{\substack{f, g \in \mathcal{C}(\mathcal{X}) \\ f \oplus g \leq c}} \int f d \mu+\int g d \nu,
\end{equation}
where the dual potentials $f, g$ are continuous functions from $\mathcal{X}$ to $\mathbb{R}$, and $f \oplus g \mapsto f(x) + g(x)$.
Based on \citet{Brenier1987}, \citet{makkuva2020optimal} derive an approximate min-max optimization scheme parameterizing the duals $f, g$ via two convex functions. The objective thereby reads
\begin{equation} \label{eq:not-minmax}
    \OT_\text{n}(\mu, \nu)=\sup _{f  \text{ convex}}\newinf _{g \text{ convex}}  \underbrace{\frac{1}{2}\mathbb{E}\left[\|x\|+\|y\|\right]}_{\mathcal{C}_{\mu, \nu}} - \underbrace{\mathbb{E}_{\mu}[f(x)]-\mathbb{E}_{\nu}[\langle y, \nabla g(y)\rangle-f(\nabla g(y))]}_{\mathcal{V}_{\mu, \nu}(f, g)}.
\end{equation}
When paramterizing $f$ and $g$ via a pair of ICNNs with parameters $\theta_f$ and $\theta_g$, this neural OT scheme then allows to predict $\nu$ or $\mu$ via $\nabla g_{\theta_g \sharp} \mu$ or $\nabla f_{\theta_f \sharp}\nu$, respectively.
We further discuss neural primal \citep{fan2021scalable, yang2018scalable} and dual approaches \citep{,makkuva2020optimal, korotin2020wasserstein, bunne2021learning} in \S\ref{sec:baselines}.

\subsection{Unbalanced optimal transport.}

A major constraint of problem \eqref{eq:ot} is its restriction to a pair of probability distributions $\mu$ and $\nu$ of equal mass.
Unbalanced optimal transport \citep{benamou2003numerical, liero2018optimal, chizat2018unbalanced} lifts this requirement and allows a comparison between unnormalized measures, i.e., via
\begin{equation} \label{eq:ub-ot}
    \inf _{\gamma \in \mathcal{M}_{+}(\gX^2)} \int_{\gX^2} c(x, y) \gamma(dx, dy) +\tau_1 \mathcal{D}_{f_1}((\Proj_1)_\sharp \gamma \mid \mu)+ \tau_2 \mathcal{D}_{f_2}((\Proj_2)_\sharp \gamma \mid \nu),
\end{equation}
with $f$-divergences $D_{f_1}$ and $D_{f_2}$ induced by $f_1, f_2$, and parameters $(\tau_1, \tau_2)$ controlling how much mass variations are penalized as opposed to transportation of the mass.
When introducing an entropy regularization as in \eqref{eq:reg-ot}, the unbalanced OT problem between discrete measures $\rvu$ and $\rvv$, i.e.,
\begin{equation} \label{eq:reg-ub-ot}
   \UBOT(\rvu, \rvv) \defeq \min _{\Gamma \in \mathbb{R}_{+}^{n \times m}}\dotp{\Gamma}{[c(x_i, y_j)]_{ij}} +\tau_1 \gD_{f_1}(\Gamma \ones_m \mid \rvu)+\tau_2 \gD_{f_2}(\Gamma^{\top} \ones_m \mid \rvv)  \,-\varepsilon H(\Gamma),
\end{equation}
can be efficiently solved via generalizations of the Sinkhorn algorithm \citep{chizat2018scaling, cuturi2013sinkhorn, benamou2015iterative}
% In the limit $\tau_1 = \tau_2 \rightarrow + \infty$ and assuming equal mass (e.g., the balanced case), one recovers \eqref{}.
.
We describe alternative formulations of the unbalanced OT problem in detail, review recent applications, and provide a broader literature review in the Appendix (\S\ref{sec:unot_add}).

\section{A Neural Unbalanced Optimal Transport Model}

The method we propose weaves together a rigorous formulation of the unbalanced optimal transport problem based on semi-couplings (introduced below) with a practical and scalable OT mapping estimation method based on input convex neural network parameterization of the dual OT problem. 

\paragraph{Semi-coupling formulation.} 
\citet{chizat2018unbalanced} introduced a class of distances that generalize optimal transport for the unbalanced setting. They introduce equivalent dynamic and static formulations of the problem, the latter of which relies on \textit{semi-couplings} to allow for variations of mass. These are generalizations of couplings whereby only one of the projections coincides with a prescribed measure. Formally, the set of semi-couplings between measures $\mu$ and $\nu$ is defined as
\begin{equation}\label{eq:semicoupling_set}
    \Gamma\left(\mu, \nu\right) \stackrel{\text { def. }}{=}\left\{\left(\gamma_0, \gamma_1\right) \in\left(\mathcal{M}_{+}\left(\gX^2\right)\right)^2:\left(\Proj_1\right)_{\sharp} \gamma_0=\mu,\left(\Proj_2\right)_{\sharp} \gamma_1=\nu\right\}.
\end{equation}
With this, the unbalanced Kantorovich OT problem can be written as $C_k(\mu, \nu) = \inf_{(\gamma_0, \gamma_1) \in  \Gamma\left(\mu, \nu\right) } \int c(x, \tfrac{\gamma_0}{\gamma}, y, \tfrac{\gamma_1}{\gamma})\dif \gamma(x,y)$, where $\gamma$ is any joint measure for which $\gamma_0, \gamma_1 \ll \gamma$.

Although this formulation lends itself to formal theoretical treatment, it has at least two limitations. First, it does not explicitly model a mapping between measures. Indeed, no analogue of the celebrated Brenier's theorem is known for this setting. Second, deriving a computational implementation of this problem is challenging by the very nature of the semi-couplings: being undetermined along one marginal makes it hard to model the space in \eqref{eq:semicoupling_set}. 

\paragraph{Rebalancing with proxy measures.} 
To turn the semi-coupling formulation of unbalanced OT into a computationally feasible method, we propose to conceptually break the problem into balanced and unbalanced subproblems, each tackling a different aspect of the difference between measures: feature transformation and mass rescaling. These in turn imply a decomposition of the semi-couplings of \eqref{eq:semicoupling_set}, as we will show later. Specifically, we seek proxy measures $\tilde{\mu}$ and $\tilde{\nu}$ with equal mass (i.e., $\mu(\gX) = \tilde{\nu}(\gX)$) across which to solve a \textit{balanced} OT problem through a Monge/Brenier formulation. To decouple measure scaling from feature transformation, we propose to choose $\tilde{\mu}$ and $\tilde{\nu}$ simply as rescaled versions of $\mu$ and $\nu$. Thus, formally, we seek $\tilde{\mu}, \tilde{\nu} \in \mathcal{M}_+(\mathcal{X})$ and $T,S: \mathcal{X} \rightarrow \mathcal{X}$ such that 
\begin{equation}\label{eq:proxy_measure_constraints}
    \tilde{\mu} = \eta \cdot \mu, \quad \tilde{\nu} = \zeta \cdot \nu, \quad T_\sharp \tilde{\mu} = \tilde{\nu}, \quad S_\sharp \tilde{\nu} = \tilde{\mu},
\end{equation}
where $\eta,\zeta: \gX \rightarrow \R^+$ are scalar fields, $\eta \cdot \mu$ denotes the measure with density $\eta(x)\dif \mu(x)$ (analogously for $\zeta \cdot \nu)$, and $T,S$ are a pair of forward/backward optimal transport maps between $\tilde{\mu}$ and $\tilde{\nu}$ (Fig.~\ref{fig:overview}b). 

Devising an optimization scheme to find all relevant components in \eqref{eq:proxy_measure_constraints} is challenging. In particular, it involves solving an OT problem whose marginals are not fixed, but that will change as the reweighting functionals $\eta,\zeta$ are updated. We propose an alternating minimization approach, whereby we alternative solve for $\eta,\zeta$ (through an approximate scaling update) and $T,S$ (through gradient updates on ICNN convex potentials, as described in Section \ref{sec:background_OT}). 

\paragraph{Updating rescaling functions.}
Given current estimates of $\eta$ and $T$, we consider the UBOT problem \eqref{eq:ub-ot} between $T_\sharp (\eta \cdot \mu) = T_\sharp \tilde{\mu}$ and $\nu$. Although in general these two measures will not be balanced (hence why we need to use UBOT instead of usual OT), our goal is to eventually achieve this. To formalize this, let us use the shorthand notation $\pi_{\textup{UB}}^*(\alpha, \beta)  \eqdef \argmin_{\pi} \textup{UBOT} (\pi;  \alpha, \beta)$, where \textup{UBOT} is defined in \eqref{eq:reg-ub-ot}. For a fixed $T$, our goal is to find $w$ such that $(\Proj_1)_\sharp [\pi_{\textup{UB}}^*(T_\sharp (\eta \cdot \mu), \nu)] = T_\sharp (\eta \cdot \mu)$. For the discrete setting (finite samples), this corresponds to finding a vector $\mathbf{e}$ satisfying:
\begin{equation}\label{eq:UBOT_fixed_point}
    \sum_{j=1}^m [\Gamma]_{ij}  = \mathbf{e} \odot \mathbf{u}, \quad \text{where} \medspace \Gamma= \argmin \textup{UBOT} (\mathbf{e} \odot \mathbf{u},T(\rvx_i), \mathbf{v}, \rvy_j) .
\end{equation}
For a fixed $T$, the vector $\rve^*$ satisfying this system can be found via a fixed-point iteration. In practice, we approximate it instead with a single-step update using the solution to the unscaled problem:
\[ \Gamma \gets  \textup{UBOT}(\mathbf{u},T(\rvx_i), \mathbf{v}, \rvy_j)  ; \qquad \mathbf{e}  \gets \Gamma \ones \oslash \mathbf{u};\]
which empirically provides a good approximation on the optimal $\rve^*$ but is significantly more efficient. Apart from requiring a single update, whenever $\mathbf{u}$ and $\mathbf{v}$ are uniform (as in most applications where the samples are assumed to be drawn i.i.d.) solving this problem between unscaled histograms will be faster and more stable than solving its scaled (and therefore likely non-uniform) counterpart in \eqref{eq:UBOT_fixed_point}.

% in closed form. In fact, Lemma~\ref{lemma:UBOT_equivalence} shows that it can be obtained by solving the  UBOT problem between the (unscaled) $T_{\sharp}\mu$ and $\nu$ instead.
% \begin{lemma}\label{lemma:UBOT_equivalence}
%     The rescaling vector $\mathbf{e}$ obtained via the update rule:
%     \[ \Gamma \gets  \textup{UBOT}(\mathbf{u},T(\rvx_i), \mathbf{v}, y_j)  ; \qquad \mathbf{e}  \gets \Gamma \ones \oslash \mathbf{u} \]
%     satisfies the marginal rescaling condition \eqref{eq:UBOT_fixed_point}.  
% \end{lemma} 
% \begin{proof}
%     ...
% \end{proof}
Analogously, for a given $S$, we choose $\zeta$ that ensures  $(\Proj_2)_\sharp [\pi_{\textup{UB}}^*(S_\sharp (\zeta \cdot \nu), \mu)] = S_\sharp (\zeta \cdot \nu)$. For empirical measures, this yields the update:
\[ \Gamma \gets  \textup{UBOT}(\mathbf{v},S(\rvy_j), \mathbf{u}, \rvx_i); \qquad \mathbf{z}  \gets \Gamma \ones \oslash \mathbf{v}; \]
In order to be able to predict mass changes for new samples, we will use the discrete $\mathbf{e}, \mathbf{z}$ to fit continuous versions of $\eta, \zeta$ via functions parameterized as neural networks trained to achieve $\eta(\rvx_i) \approx e_i \medspace \forall i \in \{1, \dots, n\}$ and $\zeta(\rvy_j) \approx z_j \medspace \forall j \in \{1, \dots, m\}$.

\paragraph{Updating mappings.}
For a fixed pair of $\eta, \zeta$, we want $T$ and $S$ to be a pair of optimal OT maps between $\tilde{\mu}$ and $\tilde{\nu}$. Since these are guaranteed to be balanced due to the argument above, we can use a usual (balanced) OT formulation to find them. In particular, we use the formulation of \citep{makkuva2020optimal} to fit them. That is, $T = \nabla g$ and $S = \nabla f$ for convex potentials $f$ and $g$, parameterized as ICNNs with parameters $\theta_f$ and $\theta_g$. The corresponding objective for these two potentials is:
\begin{align*}
    \mathcal{L}(f,g)  &= \Exp_{x\sim \tilde{\mu}} \left[ f (\nabla g) - \langle x, \nabla g (x) \rangle \right ] -  \Exp_{y\sim \tilde{\nu}} \left[ f(y) \right] \\&= \int_\mathcal{X} \bigl[ f(\nabla g (x)) - \langle x, \nabla g (x) \rangle \bigr ] \eta(x) \dif \mu(x)- \int f(y) \zeta(y) \dif \nu(y).
\end{align*}
In the finite sample setting, this objective becomes:   \vspace{-0.2cm}
\begin{equation}
    \mathcal{L}(f,g) = \frac{1}{n}\sum_{i=1}^n e_i  \left[ f (\nabla g (\rvx_i)) - \langle \rvx_i, \nabla g (\rvx_i) \rangle \right ] -  \sum_{j=1}^m z_j f(\rvy_j). \vspace{-0.2cm}
\end{equation}
The optimization procedure is summarized in Algorithm~\ref{algo}.
\vspace{-0.2cm}
\paragraph{Transforming new samples.}
After learning $f, g, \eta, \zeta$, we can use these functions to transform (map and rescale) new samples, i.e., beyond those used for optimization. For a given source datapoint $\rvx$ with mass $u$, we transform it as $(\rvx, u) \mapsto (\nabla g (\rvx), \eta(\rvx) \cdot u \cdot \zeta(\nabla g (\rvx))^{-1})$. Analogously, target points can be mapped backed to the source domain using $(\rvy, v) \mapsto (\nabla f (\rvy), \zeta(\rvy) \cdot v \cdot \eta(\nabla f (\rvy))^{-1})$.
\vspace{-0.2cm}
\paragraph{Recovering semi-couplings.}
Let us define $\tilde{\Gamma}_1 \eqdef \diag(\rve^{-1})^\top\Gamma_1$ and $\tilde{\Gamma}_2 \eqdef \diag(\rvz^{-1})^\top \Gamma_2$, where $\Gamma_1, \Gamma_2$ are the solutions of the UBOT problems computed in Algorithm 1 (lines 7 and 9, respectively). It is easy to see that $(\tilde{\Gamma}_1, \tilde{\Gamma}_2^\top)$ is a valid pair of semi-couplings between $\mu$ and $\nu$ (cf.~Eq.~\ref{eq:semicoupling_set}).

%  From the above, we can define a valid pair of semi couplings between $\mu$ and $\nu$ as follows. Let $\gamma_\mu$ be the trivial coupling between $\mu$ and $\tilde{\mu}$, i.e.,  $\gamma_\mu(x,x') = \mu(x') I(x=x')$ (in the discrete case, $\Gamma_{\mu}$ would be a diagonal matrix $\text{diag}(\mathbf{a})$). The chain together (need to formalize) $\nabla g$ and  $1/\zeta$ to get $\Gamma^*_b$ with right-marginal $\nu$. Note that this pair satisfies $(\text{Proj}_1)_\sharp \Gamma^*_a = (\text{Proj}_0)_\sharp \Gamma^*_b$, which is not necessary. We can also define a non-matching pair of couplings as follows. Take $\Gamma^*_a$ as the chaining (formalize) of $\eta$ and $\Gamma_1$, so that it has marginals $\mu$ and $(\text{Proj}_1)_\sharp \Gamma_1$. Analogously define $\Gamma^*_b$ by chaining $\Gamma_2^\top$ and $\zeta$ so that it has marginals  $(\text{Proj}_0)_\sharp \Gamma_2$ and $\nu$. 

\begin{algorithm}[t]
\KwIn{$f, g$: ICNNs, initialized s.t. $\nabla g(x) \approx x$ and $\nabla f(y) \approx y$ ; $\eta, \zeta$: NNs}
  \caption{Neural Unbalanced Optimal Transport (\textsc{NubOT})}\label{algo}
\For{t in \texttt{epochs}} {
    Sample batch $\{x_i\}_{i=1}^{n} \sim \mu$ and $\{y_j\}_{j=1}^{m} \sim \nu$

    $\hat{y} \gets \nabla g(x)$  
    
    $\hat{x} \gets \nabla f(y)$
    
    $\Gamma_1 \gets \texttt{unbalanced.sinkhorn}(\hat{y}, \tfrac{1}{n}\ones_n,  y, \tfrac{1}{m}\ones_m)$ 
    %$\tilde{w}_{\hat{y}} \gets P\texttt{.sum(axis=1)}$
    
    $e_i \gets \frac{\sum_j \Gamma_{ij}}{\sum_{ij} \Gamma_{ij}} \cdot n$ 
    
    $\Gamma_2 \gets \texttt{unbalanced.sinkhorn}(\hat{x}, \tfrac{1}{m}\ones_m,  y, \tfrac{1}{n}\ones_n)$ 
    
    %$\tilde{w}_{\hat{x}} \gets P\texttt{.sum(axis=1)}$
    
    $z_i \gets \frac{\sum_j \Gamma_{ij}}{\sum_{ij} \Gamma_{ij}} \cdot m$
    
    $J(\theta_g, \theta_f) = \frac{1}{n} \sum_{i=1}^{n} e_i \left[ f(\nabla g(x_i)) - \langle x_i, \nabla g(x_i) \rangle \right] - \frac{1}{m} \sum_{j=1}^{m} z_jf(y_j)$

    %$\hat{w}_{\hat{y}} \gets h_{fw}(x)$
    
    %$\hat{w}_{\hat{x}} \gets h_{bw}(y)$
    
    $L_{\eta}(\theta_{\eta}) = $ \texttt{MSE}$(\mathbf{e}, \eta(x))$
    
    $L_{\zeta}(\theta_{\zeta}) = $ \texttt{MSE}$(\mathbf{z}, \zeta(y))$ 
    
    Update  $\theta_g$ to minimize $J$, $\theta_{\eta}$ to minimize $L_{\eta}$, $\theta_{\zeta}$ to minimize $L_{\zeta}$, and $\theta_f$ to maximize $J$}
\end{algorithm}

\section{Evaluation} \label{sec:evaluation}
We illustrate the effectiveness of \textsc{NubOT} on different tasks, including a synthetic setup for which a ground truth matching is available, as well as an important but challenging task to predict single-cell perturbation responses to a diverse set of cancer drugs with different modes of actions.

\paragraph{Baselines.}
To put \textsc{NubOT}'s performance into perspective, we compare it to several baselines: First, we consider a balanced neural optimal transport method \textsc{CellOT} \citep{bunne2021learning},  based on the neural dual formulation of \citet{makkuva2020optimal}. Further, we benchmark \textsc{NubOT} against the current state-of-the-art \textsc{ubOT GAN}, an unbalanced OT formulation proposed by \citet{yang2018scalable}, which simultaneously learns a transport map and a scaling factor for each source point in order to account for variation of mass. Additionally, we consider two naive baselines: \textsc{Identity}, simulating the identity matching and modeling cell behavior in absence of a perturbation, and \textsc{Observed}, a random permutation of the observed target samples and thus a \emph{lower bound} when comparing predictions to observed cells on the distributional level. More details can be found in the Appendix \S\ref{sec:baselines}.

\subsection{Synthetic Data}
\begin{figure}
    \centering
    \includegraphics[width=1.05\textwidth]{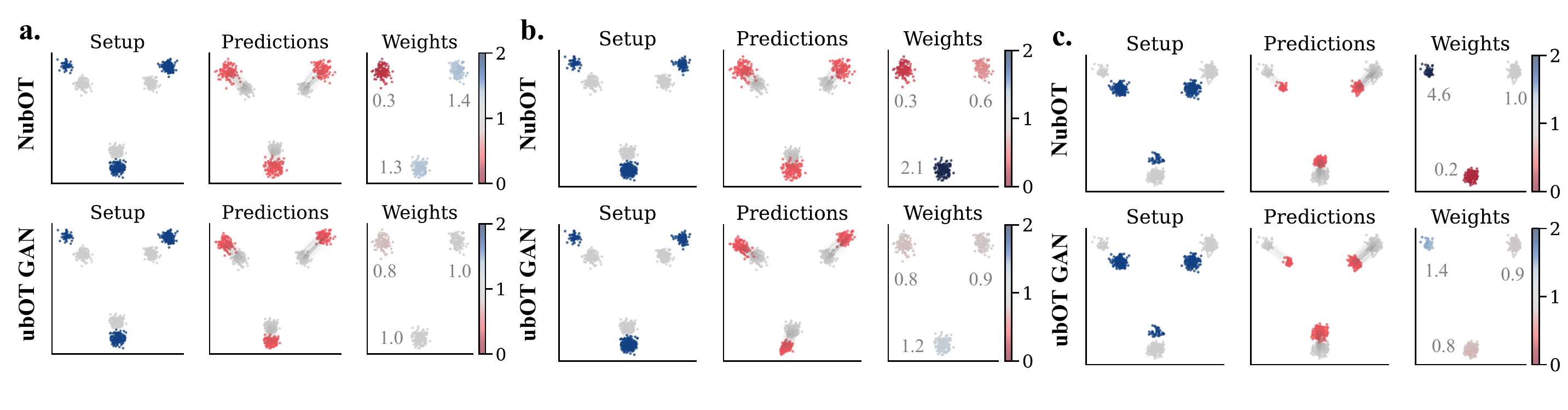}
    \caption{\looseness -1 \textbf{Unbalanced sample mapping}. In all three scenarios (a,b,c), the source (gray) and target (blue) datasets share structure but have different shifts and per-cluster sampling proportions. Tasked with mapping from source to target, \textsc{NubOT} and \textsc{ubOT GAN} predict the locations (middle pane, red) and weights (right pane) of the transported samples. The number next to the weights denotes the mean weights per cluster. While both methods map the samples to the correct location, \textsc{NubOT} more accurately predicts the weights needed to match the target distribution, creating mass (dark blue) or destroying it (red) as needed.}
    \label{fig:fig_toy}
\end{figure}

\looseness -1 Populations are often heterogeneous and consist of different subpopulations. Upon intervention, these subpopulations might exhibit heterogeneous responses. Besides a change in their feature profile, the subpopulations may also  show changes in their particle counts. To simulate such heterogeneous intervention responses, we generate a dataset containing a two-dimensional mixture of Gaussians with three clusters in the source distribution $\mu$. The target distribution $\nu$ consists of the same three clusters, but with different cluster proportions. Further, each particle has undergone a constant shift in space upon intervention. We consider three scenarios with increasing imbalance between the three clusters (see Fig.~\ref{fig:fig_toy}a-c). We evaluate \textsc{NubOT} on the task of predicting the distributional shift from source to target, while at the same time correctly rescaling the clusters such that no mass is transported across non-corresponding clusters.
\paragraph{Results.}
\looseness -1 The results (setup, predicted Monge maps and weights) are displayed in Fig.~\ref{fig:fig_toy}. Both \textsc{NubOT} and \textsc{ubOT GAN} correctly map the points to the corresponding target clusters without transporting mass across clusters. \textsc{NubOT} also accurately models the change in cluster sizes by predicting the correct weights for each point. In contrast, \textsc{ubOT GAN} only captures the general trend of cluster growth and shrinkage, but does not learn the exact weights required to re-weight the cluster proportions appropriately. The exact setup and calculation of weights can be found in the \S\ref{sec:add_exps} (see Table~\ref{tab:toy_setup}).

% Both qualitatively and quantitatively, \textsc{NubOT} outperforms all baselines by correctly re-weighting the clusters, while at the same time accurately learning to map the source sample outward to the target. 

% correctly model growth and shrinkage of clusters in varying degrees of severity / imbalance.

\subsection{Single-cell Perturbation Responses} \label{sec:exp_synth}

% Predicting responses of heterogeneous cell populations to perturbations (e.g., genetic knockouts, drugs, or developmental signals) at the level of single cells is a crucial steps towards deciphering the molecular processes underlying the studied response, which in turn is required to obtain an improved mechanistic understanding of biology in health and disease.
Through the measurement of genomic, transcriptomic, proteomic or phenotypic profiles of cells, and the identification of different cell types and cellular states based on such measurements, biologists have revealed perturbation responses and response mechanism which would have remained obscured in bulk analysis approaches \citep{green2016systems, liberali2014hierarchical, kramer2022multimodal}. However, single-cell measurements typically require the destruction the cells in the course of recording.
\begin{wrapfigure}{r}{0.34\textwidth}
  \begin{center}
    \vspace{-10pt}
    \includegraphics[width=0.4\textwidth]{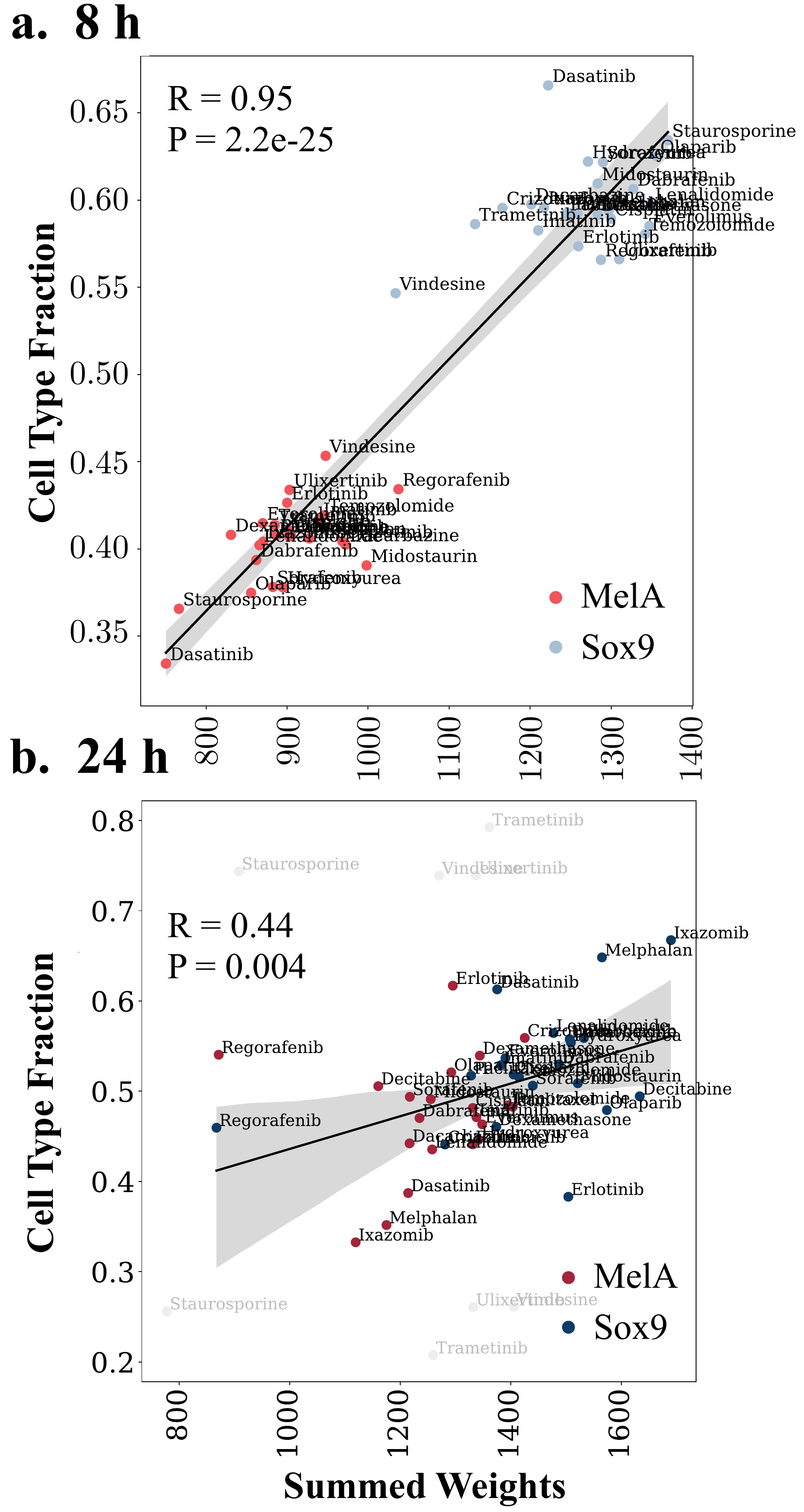}
  \end{center}
  \caption{Given the ground truth on the known subpopulation (MelA (red) and Sox9 (blue)) sizes for each drug, we analyze their level of correlation to our predicted weights after \textbf{a.} 8h and \textbf{b.} 24h. With increasing difficulty of the task and certain drugs completely removing both or one of the subpopulations, the level of correlation reduces from 8 to 24h.}
  \label{fig:summed_weights}
  \vspace{-0.75cm}
\end{wrapfigure}
% resulting in unaligned \textit{snapshots} of cell populations at different points in time.
% Tracing the development of individual cells between two measurement events (e.g., before and after treatment, from one time point to an other) remains a major challenge in computational biology, and holds the promise of drastically simplifying the identification of response mechanisms to perturbations.
\looseness -1 Thus, each measurement provides us only with a \textit{snapshot} of cell populations, i.e., samples of probability distribution that is evolving in the course of the perturbation, from control $\mu$ (source) to perturbed cell states $\nu$ (target).
Using \textsc{NubOT} and the considered baselines, we learn a map $T$ that reconstructs how individual cells respond to a treatment.
The effect of a single perturbation frequently varies depending on the cell type or cell state, and may include the induction of cell death or proliferation. In the following, we will evaluate if \textsc{NubOT} is able to capture and predict  heterogeneous proliferation and cell death rates of two co-cultures melanoma cell liens through $\eta$ and $\zeta$ in response to 25 drug treatments.
% As cells can die and proliferate in response to drug treatments, the source and target distributions are naturally unbalanced.
% Cells observed in the control population might be present in larger numbers in the treated population due to proliferation. Conversely, observed control cells might not be present at all in the treated population due to cell death. By learning the associated weights of each transported cell, we explicitly model this local variation of mass.

\begin{figure}[t]
    \centering
    \includegraphics[width=1.05\textwidth]{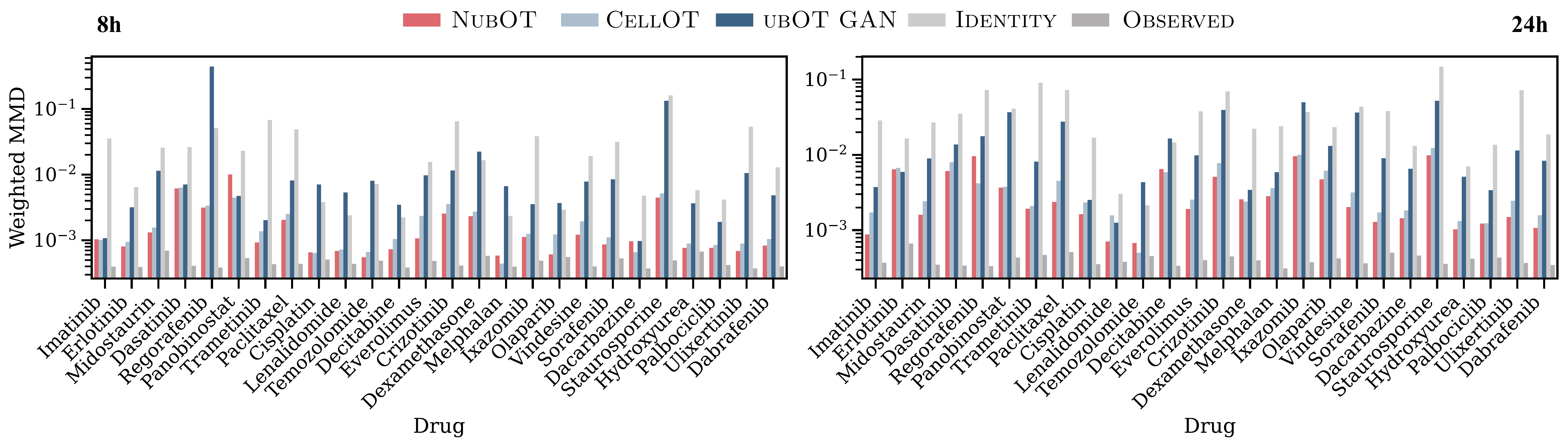}
    \caption{\looseness -1 Distributional fit of the predicted perturbed cell states to the observed perturbed cell states for each drug and timestep, measured by a weighted version of kernel MMD on a set of held-out control cells. For \textsc{NubOT} and \textsc{ubOT GAN}, MMD is weighted by the predicted weights, while for the other baselines it is computed with uniform weights. \textsc{Observed} corresponds to a random permutation of the observed control cells, i.e., its distribution is approximately the same as the observed cells.}
    \label{fig:comp_mmd_weighted}
\end{figure}

\looseness -1 The single-cell measurements used for this task were generated using the imaging technology 4i \citep{gut2018multiplexed} over the course of 24 hours, resulting in three different unaligned snapshots ($t=0h$, $t=8h$ and $t=24h$) for each of the drug treatments.
The control cells, i.e., the source distribution $\mu$, consists of cells taken from a mixture of melanoma cell lines at $t=0h$ that are exposed to a dimethyl sulfoxide (DMSO) as a vehicle control. Futher, We consider two different target populations $\nu$ capturing the perturbed populations after $t=8h$ and $t=24h$ of treatment, respectively.
As both cancer cell lines exhibit different sensitivities to the drugs \citep{raaijmakers2015new}, their proportion  (Fig.~\ref{fig:cell_counts_relative}) as well as the total cell counts (Fig.~\ref{fig:cell_counts_total}) vary over the time points.
Both cell lines are characterized by the expression of mutually exclusive protein markers, i.e., one cell line strongly expresses a set of proteins detected by an antibody called MelA (MelA$^+$ cell type), while the other is characterized by high levels of the protein Sox9 (Sox9$^+$ cell type). An evaluation of this cell line annotation can be found in Fig.~\ref{fig:cell_types_check_8h} (8h) and Fig.~\ref{fig:cell_types_check_24h} (24h).
Contrary to the synthetic task in \S~\ref{sec:exp_synth}, the nature of these measurements destroys the ground truth matching.
We thus use insights from the number of cells after 8 and 24 hours of treatment (Fig.~\ref{fig:cell_counts_relative},~\ref{fig:cell_counts_total}), as well as the cell type annotation for each cell to further evaluate \textsc{NubOT}'s performance.
A detailed description of the dataset can be found in \S~\ref{sec:cell_datasets}.

\paragraph{Results.}
\looseness -1 We split the dataset into a train and test set and train \textsc{NubOT} as well as the baselines on unaligned unperturbed (control) and perturbed cell populations for each drug. During evaluation, we then predict out-of-sample the perturbed cell state from held-out control cells. Details on the network architecture and hyperparameters can be found in \S~\ref{sec:hyperparams}.
\textsc{NubOT} and \textsc{ubOT GAN} additionally predict the weight associated with the perturbed predicted cells, giving insights into which cells have proliferated or died in response to the drug treatments.
First, we compare how well each method fits the observed perturbed cells on the level of the entire distribution. For this, we measure the weighted version of kernel maximum mean discrepancy (MMD) between predictions and observations. More details on the evaluation metrics can be found in \S~\ref{sec:eval_metrics}.
The results are displayed in Fig.~\ref{fig:comp_mmd_weighted}. \textsc{NubOT} outperforms all baselines in almost all drug perturbations, showing its effectiveness in predicting OT maps and local variation in mass.

\looseness -1 In the absence of a ground truth and in particular, given our inability to measure (i.e., observe) cells which have died upon treatment, we are required to base further analysis of \textsc{NubOT}'s predictions on changes in cell count for each subpopulation (MelA$^+$, Sox9$^+$).
Fig.~\ref{fig:cell_counts_relative} clearly shows that drug treatments lead to substantially different cell numbers for each of the subpopulations compared to control. For example, Ulixertinib leads to the proliferation of both subpopulations after 8h, but to pronounced cell death in Sox9$^+$ and strong proliferation in MelA$^+$ cells after 24h.
We thus expect, that weights predicted by \textsc{NubOT} for all drugs correlate with the change in cell counts for each cell type (here measured as population fractions).
This is indeed the case, Fig.~\ref{fig:summed_weights} shows a high correlation between observed cell counts of the two cell types and the sum of the predicted weights of the respective cell types after 8h of treatment for all drugs.
After 24 hours, treatment-induced cell death (in at least one cell type) by some drugs can be so severe at that the number of observed perturbed cells becomes too low for accurate predictions and the evaluation of the task (Fig.~\ref{fig:cell_counts_total}). Further, we find that drugs influence the abundance of the cell lines markers MelA and Sox9, complicating cell type classification (see Fig.~\ref{fig:cell_types_check_8h},~\ref{fig:cell_types_check_24h}, as well as Fig.~\ref{fig:cell_types_check_8h},~\ref{fig:cell_types_check_24h}).
We ignore drugs falling into these categories and find that whilst the correlation between predicted weights and observed cell counts is reduced after 24h (see Fig.~\ref{fig:summed_weights}a),  \textsc{NubOT} still captures the overall trend.

\begin{figure}[t]
    \centering
    \includegraphics[width=1.05\textwidth]{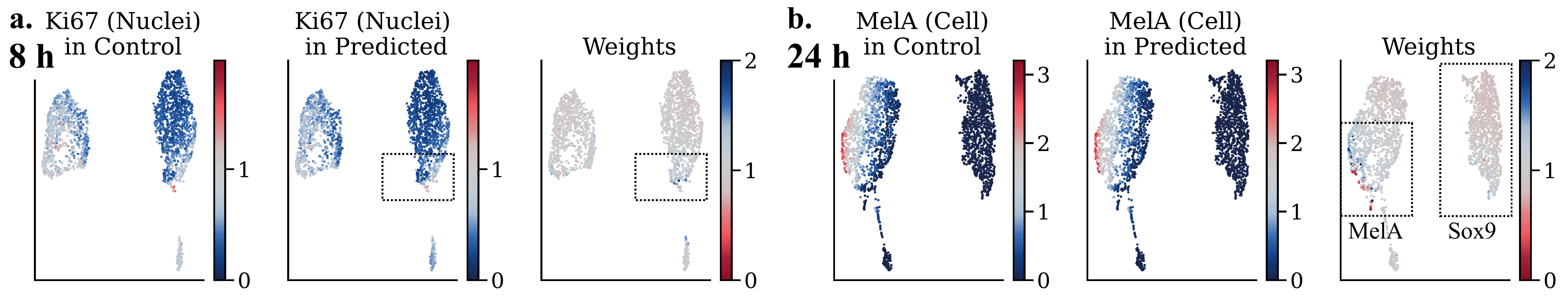}
    \caption{UMAP projections of the control cells for Ulixertinib at \textbf{a.} 8h and \textbf{b.} 24h. Cells are colored by the observed and predicted protein marker values (Ki67, MelA), and predicted weights. \textsc{NubOT} thereby correctly predicts weights $\geq 1$  for proliferating cells in the MelA$^+$ population (\textbf{a.} and \textbf{a.}, right panel), and increased levels of cell death in the Sox9$^+$ population after 24h via weights $\leq 1$ (\textbf{b.}, right panel), confirmed by the experimental observations (see Fig.~\ref{fig:cell_counts_relative}). \vspace{-15pt}}
    \label{fig:umap_ulixertinib}
\end{figure}

\looseness -1 The data further provides insights into biological processes such as apoptosis, a form of programmed cell death induced by enzymes called Caspases (ClCasp3). While dead cells become invisible in the cell state space (they cannot be measured), \textit{dying} cells are still present in the observed perturbed sample and can be recognized by high levels of ClCasp3 (the apopotosis markers).
Conversely, the protein Ki67 marks proliferating cells.
Analyzing the correlation of ClCasp3 and Ki67 intensity with the predicted weights provides an additional assessment of the biological meaningfulness of our results.
For example, upon Ulixertinib treatment, the absolute cell counts show an increase in Sox9$^+$ cells, and a decline of MelA$^+$ cells at 24h (Fig.~\ref{fig:cell_counts_relative}). Fig.~\ref{fig:umap_ulixertinib} shows UMAP projections of the control cells at both time points, colored by the observed and predicted protein marker values and the predicted weights.
At 8h, \textsc{NubOT} predicts only little change in mass, but a few proliferative cells with high weights in areas which are marked by high values of the proliferation marker Ki67. At 24h, our model predicts cell death in the Sox9$^+$ (MelA$^-$) cell type, and proliferation in the MelA$^+$ cell type, which matches the observed changes in cell counts per cell type, seen in Fig.~\ref{fig:cell_counts_relative} in \S~\ref{sec:add_exps}.
We identify similar results for Trametinib (Fig.~\ref{fig:umap_trametinib}), Ixazomib (Fig.~\ref{fig:umap_ixazomib}), and Vindesine (Fig.~\ref{fig:umap_vindesine}) which can be found in \S~\ref{sec:add_exps}.
These experiments thus demonstrate that \textsc{NubOT} accurately predicts heterogeneous drug responses at the single-cell level, capturing both, cell proliferation and death.
%\todo{Sampling proportional to observed growth rates.}

%Given access to control and treated cell populations, we predict drug responses of individual cells by learning to transport the control samples to the treated samples.

% \input{content/related} % Related work is empty now.

\section{Conclusion} \label{sec:conclusion}

\looseness -1 This work presents a novel formulation of the unbalanced optimal transport problem that bridges two previously disjoint perspectives on the topic: a theoretical one based on semi-couplings and a practical one based on recent neural estimation of OT maps. The resulting algorithm, \textsc{NubOT}, is scalable, efficient, and robust. Yet, it is effective at modeling processes that involve population growth or death, as demonstrated through various experimental results on both synthetic and real data. On the challenging single-cell perturbation task, \textsc{NubOT} is able to successfully predict perturbed cell states, while explicitly modeling death and proliferation. This is an unprecedented achievement in the field of single-cell biology, which currently relies on the use of markers to approximate the survival state of cell population upon drug treatment. Explicitly modeling proliferation and death at the single-cell level as part of the drug response, allows to link cellular properties observed prior to drug treatment to therapy outcomes. Thus, the application of \textsc{NubOT} in the fields of drug discovery and personalized medicine could be of great implications, as it allows to identify cellular properties predictive of drug efficacy.

\section*{Acknowledgments}
C.B. was supported by the NCCR Catalysis (grant number 180544), a National Centre of Competence in Research funded by the Swiss National Science Foundation. L.P. is supported by the European Research Council (ERC-2019-AdG-885579), the Swiss National Science Foundation (SNSF grant 310030\_192622), the Chan Zuckerberg Initiative, and the University of Zurich. G.G. received funding from the Swiss National Science Foundation and InnoSuisse as part of the BRIDGE program, as well as from the University of Zurich through the BioEntrepreneur Fellowship.

\bibliography{references}

\begin{thebibliography}{66}
\providecommand{\natexlab}[1]{#1}
\providecommand{\url}[1]{\texttt{#1}}
\expandafter\ifx\csname urlstyle\endcsname\relax
  \providecommand{\doi}[1]{doi: #1}\else
  \providecommand{\doi}{doi: \begingroup \urlstyle{rm}\Url}\fi

\bibitem[Alvarez-Melis et~al.(2022)Alvarez-Melis, Schiff, and
  Mroueh]{alvarez2021optimizing}
David Alvarez-Melis, Yair Schiff, and Youssef Mroueh.
\newblock {Optimizing Functionals on the Space of Probabilities with Input
  Convex Neural Networks}.
\newblock \emph{Transactions on Machine Learning Research (TMLR)}, 2022.

\bibitem[Amos et~al.(2017)Amos, Xu, and Kolter]{amos2017input}
Brandon Amos, Lei Xu, and J~Zico Kolter.
\newblock {Input Convex Neural Networks}.
\newblock In \emph{International Conference on Machine Learning (ICML)},
  volume~34, 2017.

\bibitem[Benamou(2003)]{benamou2003numerical}
Jean-David Benamou.
\newblock Numerical resolution of an “unbalanced” mass transport problem.
\newblock \emph{ESAIM: Mathematical Modeling and Numerical Analysis},
  37\penalty0 (5), 2003.

\bibitem[Benamou et~al.(2015)Benamou, Carlier, Cuturi, Nenna, and
  Peyr{\'e}]{benamou2015iterative}
Jean-David Benamou, Guillaume Carlier, Marco Cuturi, Luca Nenna, and Gabriel
  Peyr{\'e}.
\newblock {Iterative Bregman Projections for Regularized Transportation
  Problems}.
\newblock \emph{SIAM Journal on Scientific Computing}, 37\penalty0 (2), 2015.

\bibitem[Bonneel \& Coeurjolly(2019)Bonneel and Coeurjolly]{bonneel2019spot}
Nicolas Bonneel and David Coeurjolly.
\newblock {SPOT: Sliced Partial Optimal Transport}.
\newblock \emph{ACM Transactions on Graphics (TOG)}, 38\penalty0 (4), 2019.

\bibitem[Brenier(1987)]{Brenier1987}
Yann Brenier.
\newblock D{\'e}composition polaire et r{\'e}arrangement monotone des champs de
  vecteurs.
\newblock \emph{CR Acad. Sci. Paris S{\'e}r. I Math.}, 305, 1987.

\bibitem[Bunne et~al.(2021)Bunne, Stark, Gut, del Castillo, Lehmann, Pelkmans,
  Krause, and Ratsch]{bunne2021learning}
Charlotte Bunne, Stefan~G Stark, Gabriele Gut, Jacobo~Sarabia del Castillo,
  Kjong-Van Lehmann, Lucas Pelkmans, Andreas Krause, and Gunnar Ratsch.
\newblock {Learning Single-Cell Perturbation Responses using Neural Optimal
  Transport}.
\newblock \emph{bioRxiv}, 2021.

\bibitem[Bunne et~al.(2022{\natexlab{a}})Bunne, Hsieh, Cuturi, and
  Krause]{bunne2022recovering}
Charlotte Bunne, Ya-Ping Hsieh, Marci Cuturi, and Andreas Krause.
\newblock {Recovering Stochastic Dynamics via Gaussian Schr{\"o}dinger
  Bridges}.
\newblock \emph{arXiv Preprint arXiv:2202.05722}, 2022{\natexlab{a}}.

\bibitem[Bunne et~al.(2022{\natexlab{b}})Bunne, Krause, and
  Cuturi]{bunne2022supervised}
Charlotte Bunne, Andreas Krause, and Marco Cuturi.
\newblock {Supervised Training of Conditional Monge Maps}.
\newblock In \emph{Advances in Neural Information Processing Systems
  (NeurIPS)}, volume~36, 2022{\natexlab{b}}.

\bibitem[Bunne et~al.(2022{\natexlab{c}})Bunne, Meng-Papaxanthos, Krause, and
  Cuturi]{bunne2022proximal}
Charlotte Bunne, Laetitia Meng-Papaxanthos, Andreas Krause, and Marco Cuturi.
\newblock {Proximal Optimal Transport Modeling of Population Dynamics}.
\newblock In \emph{International Conference on Artificial Intelligence and
  Statistics (AISTATS)}, volume~25, 2022{\natexlab{c}}.

\bibitem[Caffarelli \& McCann(2010)Caffarelli and McCann]{caffarelli2010free}
Luis~A Caffarelli and Robert~J McCann.
\newblock {Free boundaries in optimal transport and Monge-\`Ampere obstacle
  problems}.
\newblock \emph{Annals of Mathematics}, 2010.

\bibitem[Chapel et~al.(2020)Chapel, Alaya, and Gasso]{chapel2020partial}
Laetitia Chapel, Mokhtar~Z Alaya, and Gilles Gasso.
\newblock {Partial Optimal Transport with Applications on Positive-Unlabeled
  Learning}.
\newblock In \emph{Advances in Neural Information Processing Systems
  (NeurIPS)}, volume~33, 2020.

\bibitem[Chapel et~al.(2021)Chapel, Flamary, Wu, F{\'e}votte, and
  Gasso]{chapel2021unbalanced}
Laetitia Chapel, R{\'e}mi Flamary, Haoran Wu, C{\'e}dric F{\'e}votte, and
  Gilles Gasso.
\newblock {Unbalanced Optimal Transport through Non-negative Penalized Linear
  Regression}.
\newblock \emph{Advances in Neural Information Processing Systems (NeurIPS)},
  34, 2021.

\bibitem[Chen et~al.(2019)Chen, Shi, and Zhang]{chen2018optimal}
Yize Chen, Yuanyuan Shi, and Baosen Zhang.
\newblock {Optimal Control Via Neural Networks: A Convex Approach}.
\newblock In \emph{International Conference on Learning Representations
  (ICLR)}, 2019.

\bibitem[Chizat et~al.(2018{\natexlab{a}})Chizat, Peyr{\'e}, Schmitzer, and
  Vialard]{chizat2018scaling}
Lenaic Chizat, Gabriel Peyr{\'e}, Bernhard Schmitzer, and Fran{\c{c}}ois-Xavier
  Vialard.
\newblock Scaling algorithms for unbalanced optimal transport problems.
\newblock \emph{Mathematics of Computation}, 87\penalty0 (314),
  2018{\natexlab{a}}.

\bibitem[Chizat et~al.(2018{\natexlab{b}})Chizat, Peyr{\'e}, Schmitzer, and
  Vialard]{chizat2018unbalanced}
Lenaic Chizat, Gabriel Peyr{\'e}, Bernhard Schmitzer, and Fran{\c{c}}ois-Xavier
  Vialard.
\newblock {Unbalanced optimal transport: Dynamic and Kantorovich formulations}.
\newblock \emph{Journal of Functional Analysis}, 274\penalty0 (11),
  2018{\natexlab{b}}.

\bibitem[Cuturi(2013)]{cuturi2013sinkhorn}
Marco Cuturi.
\newblock {Sinkhorn Distances: Lightspeed Computation of Optimal Transport}.
\newblock In \emph{Advances in Neural Information Processing Systems
  (NeurIPS)}, volume~26, 2013.

\bibitem[Dixit et~al.(2016)Dixit, Parnas, Li, Chen, Fulco, Jerby-Arnon,
  Marjanovic, Dionne, Burks, Raychowdhury, et~al.]{dixit2016perturb}
Atray Dixit, Oren Parnas, Biyu Li, Jenny Chen, Charles~P Fulco, Livnat
  Jerby-Arnon, Nemanja~D Marjanovic, Danielle Dionne, Tyler Burks, Raktima
  Raychowdhury, et~al.
\newblock {Perturb-Seq: Dissecting Molecular Circuits with Scalable Single-Cell
  RNA Profiling of Pooled Genetic Screens}.
\newblock \emph{Cell}, 167\penalty0 (7), 2016.

\bibitem[Dwibedi et~al.(2019)Dwibedi, Aytar, Tompson, Sermanet, and
  Zisserman]{dwibedi2019temporal}
Debidatta Dwibedi, Yusuf Aytar, Jonathan Tompson, Pierre Sermanet, and Andrew
  Zisserman.
\newblock {Temporal Cycle-Consistency Learning}.
\newblock In \emph{IEEE Conference on Computer Vision and Pattern Recognition
  (CVPR)}, 2019.

\bibitem[Fan et~al.(2021{\natexlab{a}})Fan, Liu, Ma, Chen, and
  Zhou]{fan2021scalable}
Jiaojiao Fan, Shu Liu, Shaojun Ma, Yongxin Chen, and Haomin Zhou.
\newblock {Scalable Computation of Monge Maps with General Costs}.
\newblock \emph{arXiv preprint arXiv:2106.03812}, 2021{\natexlab{a}}.

\bibitem[Fan et~al.(2021{\natexlab{b}})Fan, Taghvaei, and
  Chen]{fan2021barycenter}
Jiaojiao Fan, Amirhossein Taghvaei, and Yongxin Chen.
\newblock {Scalable Computations of Wasserstein Barycenter via Input Convex
  Neural Networks}.
\newblock In \emph{International Conference on Machine Learning (ICML)},
  2021{\natexlab{b}}.

\bibitem[Fatras et~al.(2021)Fatras, S{\'e}journ{\'e}, Flamary, and
  Courty]{fatras2021unbalanced}
Kilian Fatras, Thibault S{\'e}journ{\'e}, R{\'e}mi Flamary, and Nicolas Courty.
\newblock {Unbalanced minibatch Optimal Transport; applications to Domain
  Adaptation}.
\newblock In \emph{International Conference on Machine Learning (ICML)}, 2021.

\bibitem[Figalli(2010)]{figalli2010optimal}
Alessio Figalli.
\newblock {The Optimal Partial Transport Problem}.
\newblock \emph{Archive for Rational Mechanics and Analysis}, 195\penalty0 (2),
  2010.

\bibitem[Frogner et~al.(2015)Frogner, Zhang, Mobahi, Araya, and
  Poggio]{frogner2015learning}
Charlie Frogner, Chiyuan Zhang, Hossein Mobahi, Mauricio Araya, and Tomaso~A
  Poggio.
\newblock {Learning with a Wasserstein loss}.
\newblock In \emph{Advances in Neural Information Processing Systems
  (NeurIPS)}, volume~28, 2015.

\bibitem[Gramfort et~al.(2015)Gramfort, Peyr{\'e}, and
  Cuturi]{gramfort2015fast}
Alexandre Gramfort, Gabriel Peyr{\'e}, and Marco Cuturi.
\newblock {Fast Optimal Transport Averaging of Neuroimaging Data}.
\newblock In \emph{International Conference on Information Processing in
  Medical Imaging (IPMI)}. Springer, 2015.

\bibitem[Green \& Pelkmans(2016)Green and Pelkmans]{green2016systems}
Victoria~A Green and Lucas Pelkmans.
\newblock A systems survey of progressive host-cell reorganization during
  rotavirus infection.
\newblock \emph{Cell Host \& Microbe}, 20\penalty0 (1), 2016.

\bibitem[Gretton et~al.(2012)Gretton, Borgwardt, Rasch, Sch{\"o}lkopf, and
  Smola]{gretton2012kernel}
Arthur Gretton, Karsten~M Borgwardt, Malte~J Rasch, Bernhard Sch{\"o}lkopf, and
  Alexander Smola.
\newblock A kernel two-sample test.
\newblock \emph{Journal of Machine Learning Research}, 13, 2012.

\bibitem[Guo et~al.(2021)Guo, Jin, Wang, Qiu, Zhang, Zhu, Zhang, and
  David]{guo2021fork}
Qipeng Guo, Zhijing Jin, Ziyu Wang, Xipeng Qiu, Weinan Zhang, Jun Zhu, Zheng
  Zhang, and Wipf David.
\newblock {Fork or Fail: Cycle-Consistent Training with Many-to-One Mappings}.
\newblock In \emph{International Conference on Artificial Intelligence and
  Statistics (AISTATS)}, 2021.

\bibitem[Gut et~al.(2018)Gut, Herrmann, and Pelkmans]{gut2018multiplexed}
Gabriele Gut, Markus~D. Herrmann, and Lucas Pelkmans.
\newblock Multiplexed protein maps link subcellular organization to cellular
  states.
\newblock \emph{Science}, 361\penalty0 (6401), 2018.

\bibitem[Hellinger(1909)]{hellinger1909neue}
Ernst Hellinger.
\newblock {Neue Begr{\"u}ndung der Theorie quadratischer Formen von
  unendlichvielen Ver{\"a}nderlichen}.
\newblock \emph{Journal f{\"u}r die reine und angewandte Mathematik},
  1909\penalty0 (136), 1909.

\bibitem[Hoffman et~al.(2018)Hoffman, Tzeng, Park, Zhu, Isola, Saenko, Efros,
  and Darrell]{hoffman2018cycada}
Judy Hoffman, Eric Tzeng, Taesung Park, Jun-Yan Zhu, Phillip Isola, Kate
  Saenko, Alexei Efros, and Trevor Darrell.
\newblock {CyCADA: Cycle-Consistent Adversarial Domain Adaptation}.
\newblock In \emph{International Conference on Machine Learning (ICML)}, 2018.

\bibitem[Huang et~al.(2021)Huang, Chen, Tsirigotis, and
  Courville]{huang2021convex}
Chin-Wei Huang, Ricky T.~Q. Chen, Christos Tsirigotis, and Aaron Courville.
\newblock {Convex Potential Flows: Universal Probability Distributions with
  Optimal Transport and Convex Optimization}.
\newblock In \emph{International Conference on Learning Representations
  (ICLR)}, 2021.

\bibitem[Hur et~al.(2021)Hur, Guo, and Liang]{hur2021reversible}
YoonHaeng Hur, Wenxuan Guo, and Tengyuan Liang.
\newblock {Reversible Gromov-Monge Sampler for Simulation-Based Inference}.
\newblock \emph{arXiv preprint arXiv:2109.14090}, 2021.

\bibitem[Janati et~al.(2020)Janati, Cuturi, and Gramfort]{janati2020spatio}
Hicham Janati, Marco Cuturi, and Alexandre Gramfort.
\newblock {Spatio-Temporal Alignments: Optimal transport through space and
  time}.
\newblock In \emph{International Conference on Artificial Intelligence and
  Statistics (AISTATS)}, volume~23, 2020.

\bibitem[Kalal et~al.(2010)Kalal, Mikolajczyk, and Matas]{kalal2010forward}
Zdenek Kalal, Krystian Mikolajczyk, and Jiri Matas.
\newblock {Forward-Backward Error: Automatic Detection of Tracking Failures}.
\newblock In \emph{IEEE Conference on Computer Vision and Pattern Recognition
  (CVPR)}. IEEE, 2010.

\bibitem[Kantorovich(1942)]{kantorovich1942transfer}
L~Kantorovich.
\newblock {On the transfer of masses (in Russian)}.
\newblock In \emph{Doklady Akademii Nauk}, volume~37, 1942.

\bibitem[Korotin et~al.(2020)Korotin, Egiazarian, Asadulaev, Safin, and
  Burnaev]{korotin2020wasserstein}
Alexander Korotin, Vage Egiazarian, Arip Asadulaev, Alexander Safin, and Evgeny
  Burnaev.
\newblock Wasserstein-2 generative networks.
\newblock In \emph{International Conference on Learning Representations}, 2020.

\bibitem[Korotin et~al.(2021)Korotin, Li, Genevay, Solomon, Filippov, and
  Burnaev]{korotin2021neural}
Alexander Korotin, Lingxiao Li, Aude Genevay, Justin~M Solomon, Alexander
  Filippov, and Evgeny Burnaev.
\newblock Do neural optimal transport solvers work? a continuous wasserstein-2
  benchmark.
\newblock \emph{Advances in Neural Information Processing Systems}, 34, 2021.

\bibitem[Kramer et~al.(2022)Kramer, Sarabia~del Castillo, and
  Pelkmans]{kramer2022multimodal}
Bernhard~A Kramer, Jacobo Sarabia~del Castillo, and Lucas Pelkmans.
\newblock Multimodal perception links cellular state to decision-making in
  single cells.
\newblock \emph{Science}, 377\penalty0 (6606):\penalty0 642--648, 2022.

\bibitem[Lee et~al.(2019)Lee, Bertrand, and Rozell]{lee2019parallel}
John Lee, Nicholas~P Bertrand, and Christopher~J Rozell.
\newblock {Parallel Unbalanced Optimal Transport Regularization for Large Scale
  Imaging Problems}.
\newblock \emph{arXiv preprint arXiv:1909.00149}, 2019.

\bibitem[Liberali et~al.(2014)Liberali, Snijder, and
  Pelkmans]{liberali2014hierarchical}
Prisca Liberali, Berend Snijder, and Lucas Pelkmans.
\newblock A hierarchical map of regulatory genetic interactions in membrane
  trafficking.
\newblock \emph{Cell}, 157\penalty0 (6):\penalty0 1473--1487, 2014.

\bibitem[Liero et~al.(2018)Liero, Mielke, and Savar{\'e}]{liero2018optimal}
Matthias Liero, Alexander Mielke, and Giuseppe Savar{\'e}.
\newblock {Optimal Entropy-Transport problems and a new Hellinger–Kantorovich
  distance between positive measures}.
\newblock \emph{Inventiones Mathematicae}, 211\penalty0 (3), 2018.

\bibitem[Lotfollahi et~al.(2019)Lotfollahi, Wolf, and
  Theis]{lotfollahi2019scgen}
Mohammad Lotfollahi, F~Alexander Wolf, and Fabian~J Theis.
\newblock {scGen predicts single-cell perturbation responses}.
\newblock \emph{Nature Methods}, 16\penalty0 (8), 2019.

\bibitem[Makkuva et~al.(2020)Makkuva, Taghvaei, Oh, and
  Lee]{makkuva2020optimal}
Ashok Makkuva, Amirhossein Taghvaei, Sewoong Oh, and Jason Lee.
\newblock Optimal transport mapping via input convex neural networks.
\newblock In \emph{International Conference on Machine Learning (ICML)},
  volume~37, 2020.

\bibitem[M{\'e}moli \& Needham(2022)M{\'e}moli and Needham]{memoli2018distance}
Facundo M{\'e}moli and Tom Needham.
\newblock Distance distributions and inverse problems for metric measure
  spaces.
\newblock \emph{Studies in Applied Mathematics}, 2022.

\bibitem[Monge(1781)]{Monge1781}
Gaspard Monge.
\newblock M{\'e}moire sur la th{\'e}orie des d{\'e}blais et des remblais.
\newblock \emph{Histoire de l'Acad{\'e}mie Royale des Sciences}, pp.\
  666--704, 1781.

\bibitem[Pele \& Werman(2009)Pele and Werman]{pele2009fast}
Ofir Pele and Michael Werman.
\newblock {Fast and robust Earth Mover's Distances}.
\newblock In \emph{International Conference on Computer Vision (ICCV)}. IEEE,
  2009.

\bibitem[Peyré \& Cuturi(2019)Peyré and Cuturi]{Peyre2019computational}
Gabriel Peyré and Marco Cuturi.
\newblock {Computational Optimal Transport}.
\newblock \emph{Foundations and Trends in Machine Learning}, 11\penalty0 (5-6),
  2019.
\newblock ISSN 1935-8245.

\bibitem[Pham et~al.(2020)Pham, Le, Ho, Pham, and Bui]{pham2020unbalanced}
Khiem Pham, Khang Le, Nhat Ho, Tung Pham, and Hung Bui.
\newblock {On Unbalanced Optimal Transport: An Analysis of Sinkhorn Algorithm}.
\newblock In \emph{International Conference on Machine Learning (ICML)}, 2020.

\bibitem[Raaijmakers et~al.(2015)Raaijmakers, Widmer, Maudrich, Koch, Langer,
  Flace, Schnyder, Dummer, and Levesque]{raaijmakers2015new}
Marieke~IG Raaijmakers, Daniel~S Widmer, Melanie Maudrich, Tabea Koch, Alice
  Langer, Anna Flace, Claudia Schnyder, Reinhard Dummer, and Mitchell~P
  Levesque.
\newblock A new live-cell biobank workflow efficiently recovers heterogeneous
  melanoma cells from native biopsies.
\newblock \emph{Experimental Dermatology}, 24\penalty0 (5):\penalty0 377--380,
  2015.

\bibitem[Richter-Powell et~al.(2021)Richter-Powell, Lorraine, and
  Amos]{richter2021input}
Jack Richter-Powell, Jonathan Lorraine, and Brandon Amos.
\newblock Input convex gradient networks.
\newblock \emph{arXiv preprint arXiv:2111.12187}, 2021.

\bibitem[Santambrogio(2015)]{santambrogio2015optimal}
Filippo Santambrogio.
\newblock {Optimal Transport for Applied Mathematicians}.
\newblock \emph{Birkh{\"a}user, NY}, 55\penalty0 (58-63):\penalty0 94, 2015.

\bibitem[Sarlin et~al.(2020)Sarlin, DeTone, Malisiewicz, and
  Rabinovich]{sarlin2020superglue}
Paul-Edouard Sarlin, Daniel DeTone, Tomasz Malisiewicz, and Andrew Rabinovich.
\newblock {SuperGlue: Learning Feature Matching with Graph Neural Networks}.
\newblock In \emph{IEEE Conference on Computer Vision and Pattern Recognition
  (CVPR)}, 2020.

\bibitem[Schiebinger et~al.(2019)Schiebinger, Shu, Tabaka, Cleary, Subramanian,
  Solomon, Gould, Liu, Lin, Berube, et~al.]{schiebinger2019optimal}
Geoffrey Schiebinger, Jian Shu, Marcin Tabaka, Brian Cleary, Vidya Subramanian,
  Aryeh Solomon, Joshua Gould, Siyan Liu, Stacie Lin, Peter Berube, et~al.
\newblock {Optimal-Transport Analysis of Single-Cell Gene Expression Identifies
  Developmental Trajectories in Reprogramming}.
\newblock \emph{Cell}, 176\penalty0 (4), 2019.

\bibitem[S{\'e}journ{\'e} et~al.(2022)S{\'e}journ{\'e}, Vialard, and
  Peyr{\'e}]{sejourne2022faster}
Thibault S{\'e}journ{\'e}, Fran{\c{c}}ois-Xavier Vialard, and Gabriel
  Peyr{\'e}.
\newblock {Faster Unbalanced Optimal Transport: Translation invariant Sinkhorn
  and 1-D Frank-Wolfe}.
\newblock In \emph{International Conference on Artificial Intelligence and
  Statistics (AISTATS)}, 2022.

\bibitem[Sheldon et~al.(2007)Sheldon, Elmohamed, and
  Kozen]{sheldon2007collective}
Daniel Sheldon, MA~Elmohamed, and Dexter Kozen.
\newblock {Collective Inference on Markov Models for Modeling Bird Migration}.
\newblock In \emph{Advances in Neural Information Processing Systems
  (NeurIPS)}, volume~20, 2007.

\bibitem[Shen et~al.(2017)Shen, Lei, Barzilay, and Jaakkola]{shen2017style}
Tianxiao Shen, Tao Lei, Regina Barzilay, and Tommi Jaakkola.
\newblock {Style Transfer from Non-Parallel Text by Cross-Alignment}.
\newblock \emph{Advances in Neural Information Processing Systems (NeurIPS)},
  30, 2017.

\bibitem[Tong et~al.(2020)Tong, Huang, Wolf, Van~Dijk, and
  Krishnaswamy]{tong2020trajectorynet}
Alexander Tong, Jessie Huang, Guy Wolf, David Van~Dijk, and Smita Krishnaswamy.
\newblock {TrajectoryNet: A Dynamic Optimal Transport Network for Modeling
  Cellular Dynamics}.
\newblock In \emph{International Conference on Machine Learning (ICML)}, 2020.

\bibitem[Van~der Walt et~al.(2014)Van~der Walt, Sch{\"o}nberger,
  Nunez-Iglesias, Boulogne, Warner, Yager, Gouillart, and Yu]{van2014scikit}
Stefan Van~der Walt, Johannes~L Sch{\"o}nberger, Juan Nunez-Iglesias,
  Fran{\c{c}}ois Boulogne, Joshua~D Warner, Neil Yager, Emmanuelle Gouillart,
  and Tony Yu.
\newblock scikit-image: image processing in python.
\newblock \emph{PeerJ}, 2:\penalty0 e453, 2014.

\bibitem[Yang \& Uhler(2019)Yang and Uhler]{yang2018scalable}
Karren~D Yang and Caroline Uhler.
\newblock {Scalable Unbalanced Optimal Transport using Generative Adversarial
  Networks}.
\newblock \emph{International Conference on Learning Representations (ICLR)},
  2019.

\bibitem[Yuan et~al.(2021)Yuan, Shen, Luna, Korkut, Marks, Ingraham, and
  Sander]{yuan2021cellbox}
Bo~Yuan, Ciyue Shen, Augustin Luna, Anil Korkut, Debora~S Marks, John Ingraham,
  and Chris Sander.
\newblock {CellBox: interpretable machine learning for perturbation biology
  with application to the design of cancer combination therapy}.
\newblock \emph{Cell Systems}, 12\penalty0 (2), 2021.

\bibitem[Zhan et~al.(2019)Zhan, Ambrosi, Wandmacher, Rauscher, Betge,
  Rindtorff, H{\"a}ussler, Hinsenkamp, Bamberg, Hessling, et~al.]{zhan2019mek}
Tianzuo Zhan, Giulia Ambrosi, Anna~Maxi Wandmacher, Benedikt Rauscher, Johannes
  Betge, Niklas Rindtorff, Ragna~S H{\"a}ussler, Isabel Hinsenkamp, Leonhard
  Bamberg, Bernd Hessling, et~al.
\newblock {MEK inhibitors activate Wnt signalling and induce stem cell
  plasticity in colorectal cancer}.
\newblock \emph{Nature Communications}, 10\penalty0 (1), 2019.

\bibitem[Zhang et~al.(2021)Zhang, Xiao, Efros, Pinto, and
  Wang]{zhang2020learning}
Qiang Zhang, Tete Xiao, Alexei~A Efros, Lerrel Pinto, and Xiaolong Wang.
\newblock {Learning Cross-Domain Correspondence for Control with Dynamics
  Cycle-Consistency}.
\newblock \emph{International Conference on Learning Representations (ICLR)},
  2021.

\bibitem[Zhang et~al.(2022)Zhang, Mroueh, Goldfeld, and
  Sriperumbudur]{zhang2022cycle}
Zhengxin Zhang, Youssef Mroueh, Ziv Goldfeld, and Bharath Sriperumbudur.
\newblock {Cycle Consistent Probability Divergences Across Different Spaces}.
\newblock In \emph{International Conference on Artificial Intelligence and
  Statistics (AISTATS)}, 2022.

\bibitem[Zhu et~al.(2017{\natexlab{a}})Zhu, Park, Isola, and
  Efros]{zhu2017unpaired}
Jun-Yan Zhu, Taesung Park, Phillip Isola, and Alexei~A Efros.
\newblock {Unpaired Image-to-Image Translation using Cycle-Consistent
  Adversarial Networks}.
\newblock In \emph{IEEE Conference on Computer Vision and Pattern Recognition
  (CVPR)}, 2017{\natexlab{a}}.

\bibitem[Zhu et~al.(2017{\natexlab{b}})Zhu, Zhang, Pathak, Darrell, Efros,
  Wang, and Shechtman]{zhu2017toward}
Jun-Yan Zhu, Richard Zhang, Deepak Pathak, Trevor Darrell, Alexei~A Efros,
  Oliver Wang, and Eli Shechtman.
\newblock {Toward Multimodal Image-to-Image Translation}.
\newblock In \emph{30}, 2017{\natexlab{b}}.

\end{thebibliography}
\bibliographystyle{iclr2023_conference}

\clearpage
\newpage

\appendix
{\LARGE\sc Appendix \par}

\section{Background} \label{sec:ext_background}
In the following, we provide further information and review related literature on concepts discussed throughout this work.

\subsection{Unbalanced Optimal Transport} \label{sec:unot_add}

\looseness -1 Unbalanced optimal transport is a generalization of the classical OT formulation \eqref{eq:ot}, and as such allows mass to be created and destroyed throughout the transport. This relaxation has found recent use cases in various domains ranging from biology \citep{schiebinger2019optimal, yang2018scalable}, imaging \citep{lee2019parallel}, shape registration \citep{bonneel2019spot}, domain adaption \citep{fatras2021unbalanced}, positive-unlabeled learning \citep{chapel2020partial}, to general machine learning \citep{janati2020spatio, frogner2015learning}.
Problem \eqref{eq:ub-ot} provides a general framework of the unbalanced optimal transport problem, which can recover related notions introduced in the literature:
Choosing for $\gD_f$ the Kullback-Leibler divergence, one recovers the so-called squared \citeauthor{hellinger1909neue} distance. Alternatively, with $\gD_f = \ell_2$ norm, we arrive at \citet{benamou2003numerical}, while an $\ell_1$ norm retrieves a concept often referred to as partial OT \citep{figalli2010optimal}. The latter comprises approaches which do not rely on a relaxation of the marginal constraints as in \eqref{eq:ub-ot}. In particular, some strategies of partial OT expand the original problem by adding \emph{virtual} mass to the marginals \citep{pele2009fast, caffarelli2010free, gramfort2015fast}, or by extending the OT map by \emph{dummy} rows and columns \citep{sarlin2020superglue} onto which excess mass can be transported.
A further review is provided in \citep[Chapter 10.2]{Peyre2019computational}.
Recent work has furthermore developed alternative computational schemes \citep{chapel2021unbalanced, sejourne2022faster} as well as provided a computational complexity analysis \citep{pham2020unbalanced} of the generalized Sinkhorn algorithm solving entropic regularized unbalanced OT \citep{chizat2018scaling}. Besides \citet{yang2018scalable}, these approaches do not provide parameterizations of the unbalanced problem and allow for an out-of-sample generalization which we consider in this work.

\subsection{Cycle-Consistent Learning} \label{sec:cyclecon}

The principle of cycle-consistency has been widely used for learning bi-directional transformations between two domains of interest. Cycle-consistency thereby assumes that both the forward and backward mapping are roughly inverses of each other.
In particular, given unaligned points $x \in \gX$ and $y \in \gY$, as well as maps $g : \gX \mapsto \gY$ and $f : \gY \mapsto \gX$, cycle-consistency reconstruction losses enforce $\| x - f(g(x)) \|$ as well as $\| y - g(f(y)) \|$ using some notion of  distance $\| \cdot \|$, assuming that there exists such a ground truth bijection $g = f^{-1}$ and $f = g^{-1}$.
The advantage of validating \emph{good} matches by cycling between \emph{unpaired} samples becomes evident through the numerous use cases to which cycle-consistency has been applied: Originally introduced within the field of computer vision \citep{kalal2010forward} and applied to image-to-image translation tasks \citep{zhu2017unpaired}, it has been quickly adapted to multi-modal problems \citep{zhu2017toward}, domain adaptation \citep{hoffman2018cycada}, and natural language processing \citep{shen2017style}.
The original principle has been further generalized to settings requiring a many-to-one or surjective mapping between domains \citep{guo2021fork} via conditional variational autoencoders, dynamic notions of cycle-consistency \citep{zhang2020learning}, or to time-varying applications \citep{dwibedi2019temporal}.
These classical approaches enforce cycle-consistency by \emph{explicitly} composing both maps and penalizing for any deviation from this bijection. In this work, we treat cycle-consistency differently. It is enforced implicitly by coupling the two distributions of interest through a sequence of reversible transformations: re-weighting, transforming, and re-weighting (Eq.~\eqref{eq:proxy_measure_constraints} and Fig.~\ref{fig:overview}).

Similarly to our work, \citet{zhang2022cycle} and \citet{hur2021reversible} establish a notion of cycle-consistency (reversibility) for a pair of pushforward operators to align two unpaired measures.
Both methods rely on the Gromov-Monge distance \citep{memoli2018distance}, a divergence to compare probability distributions defined on different ambient spaces $\gX$ and $\gY$---a setting not considered in this work.
They proceed by defining a reversible metric through replacing the single Monge map by a pair of two Monge maps, i.e., $f: \gX \mapsto \gY$ and $g: \gY \mapsto \gX$, minimizing the objective
\begin{equation} \label{eq:gromov-monge}
    \mathrm{GM}(\mu, \nu):=\inf _{\substack{f: \gX \mapsto \gY, f_{\sharp} \mu=\nu \\ g: \gY \mapsto \gX, g_{\sharp} \nu=\mu}} \Delta_{\gX}^{p}(f ; \mu)+\Delta_{\gY}^{p}(g ; \nu)+ \Delta_{\gX, \gY}^{p}(f, g ; \mu, \nu),
\end{equation}
\begin{align*}
    \Delta_{\gX}^{p}(f ; \mu) &= \left( \mathbb{E} \left[ | c_\gX(x, x') - c_\gY(f(x), f(x')) |^p \right] \right)^\frac{1}{p} \\
    \Delta_{\gY}^{p}(g ; \nu) &= \left( \mathbb{E} \left[ | c_\gX(y, y') - c_\gY(g(y), g(y')) |^p \right] \right)^\frac{1}{p} \\
    \Delta_{\gX, \gY}^{p}(f, g ; \mu, \nu) &= \left( \mathbb{E} \left[ | c_\gX(x, g(y)) - c_\gY(f(x), y) |^p \right] \right)^\frac{1}{p}.
\end{align*}
Problem \eqref{eq:gromov-monge} shows similarities to the classical cycle-consistency objective of \citet{zhu2017unpaired}, where cycle-consistency is indirectly enforced through $\Delta_{\gX, \gY}^{p}$. \citet{zhang2022cycle} parameterize both Monge maps through neural networks in a similar fashion as done in \citep{yang2018scalable, fan2021scalable}.
Our approach differs from \citet{zhang2022cycle, hur2021reversible} as we model the problem through a single Monge map with duals $f, g$, allowing us to map back-and-forth between measures $\mu$ and $\nu$, and using a different parametrization approach (ICNNs). More importantly, the approaches presented by \citet{zhang2022cycle, hur2021reversible} do not generalize to the unbalanced case. While \citet{zhang2022cycle} proposed an unbalanced version of \eqref{eq:gromov-monge} by relaxing the marginals as done in \citet{chizat2018scaling}, they require the unbalanced sample sizes to be known (i.e., $n$ and $m$ need to be fixed). In our application of interest, particle counts of the target population are, however, not known \emph{a priori}.

\subsection{Convex Neural Architectures} \label{sec:icnn}

Input convex neural networks \citep{amos2017input} are a class of neural networks that approximate the family of convex functions $\psi$ with parameters $\theta$, i.e., whose outputs $\psi_{\theta}(x)$ are convex w.r.t. an input $x$. This property is realized by placing certain constraints on the networks parameters $\theta$. More specifically, an ICNN is an $L$-layer feed-forward neural network, where each layer $l = \{0, ... ,L-1\}$ is given by 
\begin{equation} \label{eq:icnn}
    z_{l+1} = \sigma_l(W^x_l x + W^z_l z_l + b_l)  \text{ and } \psi_\theta(x) = z_L,
\end{equation}
where $\sigma_k$ are convex non-decreasing activation functions, and $\theta = \{W^x_l, W^z_l, b_l\}_{l=0}^{L-1}$ is the set of parameters, with all entries in $W^z_l$ being non-negative and the convention that $z_0$ and $W^z_0$ are $0$. 
As mentioned above and through the connection established in \S~\ref{sec:background}, convex neural networks have been utilized to approximate Monge map $T$ \eqref{eq:monge} via the convex Brenier potential $\psi$ connected to the primal and
dual optimal transport problem. In particular, it has been used to model convex dual functions \citep{makkuva2020optimal} as well as normalizing flows derived from convex potentials \citep{huang2021convex}. The expressivity and universal approximation properties of ICNNs have been further studied by~\citet{chen2018optimal}, who show that any convex function over a compact convex domain can be approximated in sup norm by an ICNN.
To improve convergence and robustness of ICNNs ---known to be notoriously difficult to train \citep{richter2021input}--- different initialization schemes have been proposed: \citet{bunne2022supervised} derive two initialization schemes ensuring that \emph{upon initialization} $\nabla \psi$  mimics an affine Monge map $T$ mapping either the source measure onto itself (identity initialization) or providing a map between Gaussian approximations of measures $\mu$ and $\nu$ (Gaussian initialization). Further, \citet{korotin2020wasserstein} proposed to use quadratic layers as well as a pre-training pipeline to initialize ICNN parameters to encode an identity map.

\section{Additional Experimental Results} \label{sec:add_exps}

\subsection{Synthetic Data}

\begin{figure}[h]
    \centering
    \includegraphics[width=0.45\textwidth]{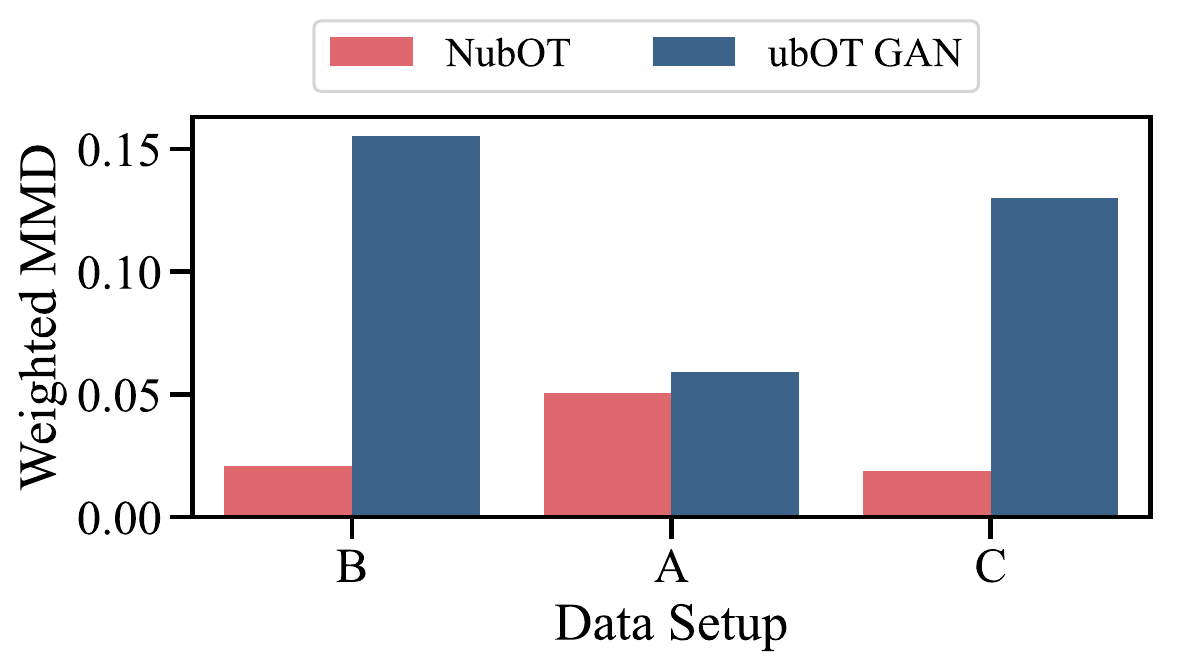}
    \caption{Distributional fit of the predicted samples to the target samples on synthetic data, measured by a weighted version of kernel MMD.}
    \label{fig:toy_mmd_w}
\end{figure}

In our synthetic two-dimensional dataset, the source and target distribution are mixtures of Gaussians with varying proportions (see Fig. \ref{fig:fig_toy}). Both source and target consist of three corresponding clusters, and by changing the proportions of each cluster, we illustrate a scenario in which subpopulations grow and shrink at different rates. Table~\ref{tab:toy_setup} shows the shares of the three clusters in the source and target distributions. In order to match the target distribution without transporting mass across non-corresponding clusters, the clusters have to be re-scaled with the factors presented in column 'True Scaling Factor'. The last two columns show the mean weights per cluster obtained by \textsc{NubOT} and \textsc{ubOT GAN}, respectively. \textsc{ubOT GAN} captures only the general trend in growth and shrinkage, the exact weights do not scale the cluster proportions appropriately. In contrast, the weights obtained by \textsc{NubOT} match the required scaling factors very closely. Fig. \ref{fig:toy_mmd_w}, shows the weighted MMD between the source distribution and the target distribution, confirming superior performance of \textsc{NubOT}.

\begin{table}[h]
    \centering
    \caption{Setup of the synthetic mixture of Gaussians dataset, showing the proportions of the three clusters in source and target distribution in three different settings (\textbf{a.}, \textbf{b.}, \textbf{c.}) as well as the required scaling factor per cluster needed to match the target without transporting points to non-corresponding clusters. The last two columns show the mean weights obtained by \textsc{NubOT} and \textsc{ubOT GAN}.}
    \label{tab:toy_setup}
    \footnotesize
    \begin{tabular}{p{0.3cm}|p{1cm}p{2cm}p{2cm}p{2cm}|p{2cm}p{2cm}}
    \toprule & \textbf{Cluster} &
    \textbf{Source Proportions} ($p$) & \textbf{Target Proportions} ($q$) & \textbf{True Scaling Factor ($q/p$)} & \textbf{Mean Weight \textsc{NubOT}} & \textbf{Mean Weight \textsc{ubOT GAN}} \\
    \midrule
    \textbf{a.} &1  & 0.33  & 0.45 & 1.35   & 1.32 & 1.02 \\
    & 2&0.33 & 0.45 & 1.35  & 1.36& 0.99 \\
    & 3&0.33 & 0.10 & 0.30 & 0.26 & 0.8\\
    \midrule
    \textbf{b.} &1& 0.33 & 0.70 & 2.10 & 2.07 & 1.18 \\
    &2& 0.33 & 0.20 & 0.60 & 0.64 & 0.88 \\
    &3& 0.33 & 0.10 & 0.30 & 0.29 & 0.81 \\
    \midrule
    \textbf{c.}&1 & 0.45 & 0.10 & 0.22 & 0.23 & 0.79 \\
    &2& 0.45 & 0.45 & 1.00 & 0.98 & 0.94 \\
    &3& 0.10 & 0.45 & 4.50 & 4.60 & 1.44 \\
    \bottomrule
    \end{tabular}
\end{table}

\subsection{Single-Cell Perturbation Responses}

\begin{figure}
    \centering
    \includegraphics[width=0.6\textwidth]{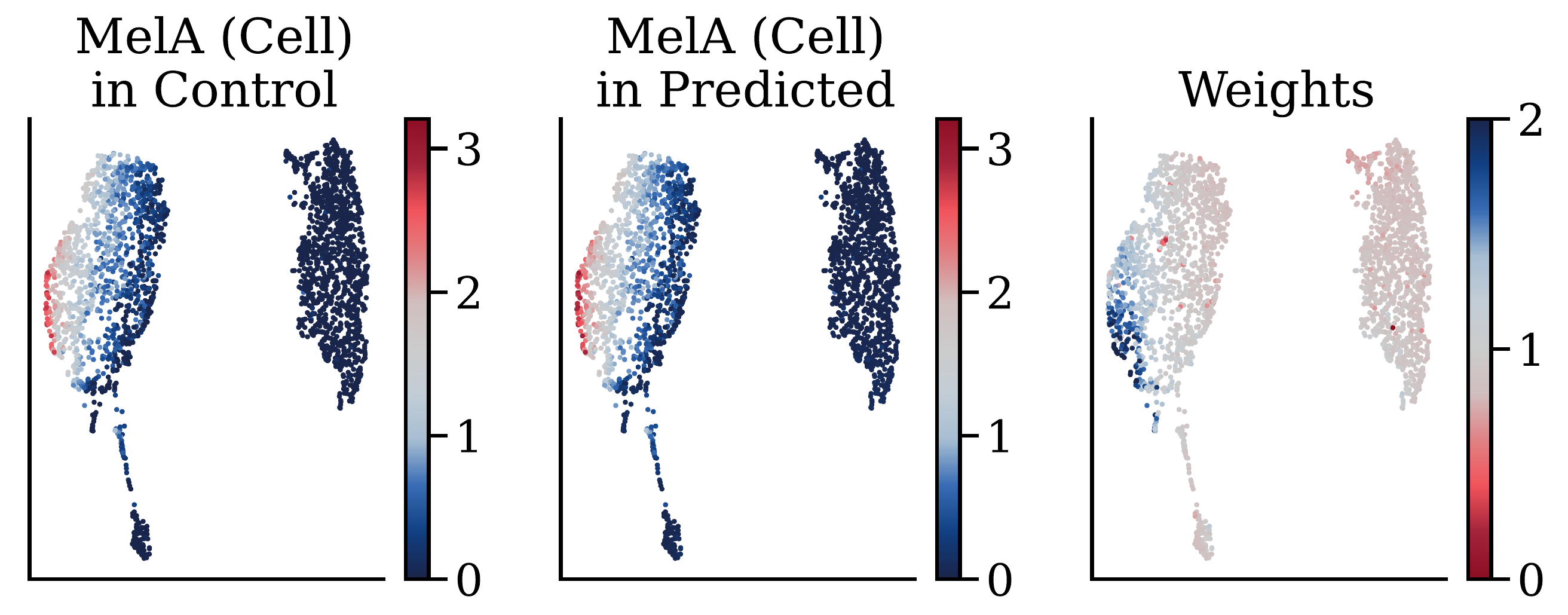}
    \caption{UMAP projections computed on control cells for Trametinib at $t=24h$. High predicted weights in the MelA$^+$ cell type suggest proliferation, while the Sox9$^+$ population shows higher levels of cell death. This prediction is confirmed by the relative cell counts, where MelA$^+$ cell counts increase and Sox9$^+$ counts decrease, demonstrating opposite response behaviors to Trametinib for each subpopulation, i.e., MelA$^+$ cells show proliferation and Sox9$^+$ cells death.}
    \label{fig:umap_trametinib}
\end{figure}

\begin{figure}
    \centering
    \includegraphics[width=\textwidth]{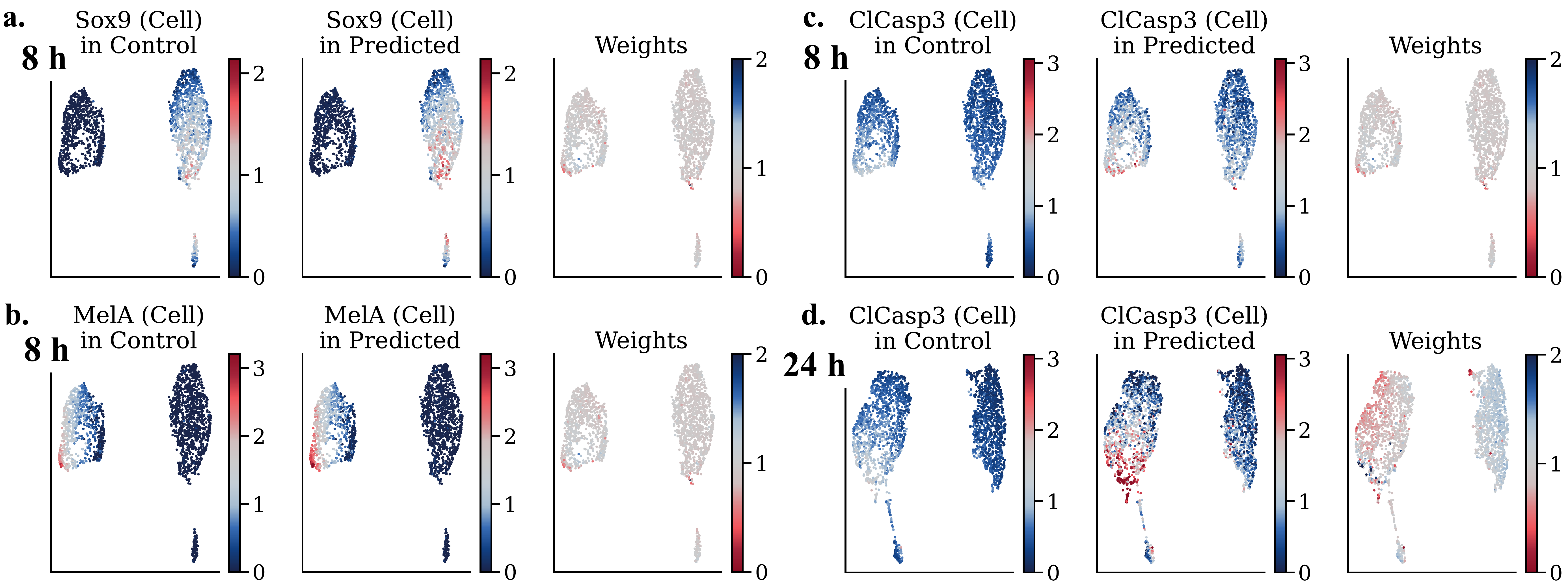}
    \caption{UMAP projections computed on control cells for Ixazomib for $t=8h$, and $t=24h$, colored by protein marker intensities \textbf{a.} MelA and \textbf{b.} Sox9, markers for the two subpopulations, as well as ClCasp3, a marker for cell death, at \textbf{c.} 8h and \textbf{d.} 24h. The UMAPs confirm the measured relative cell counts of each subpopulation. After 8h \textbf{a.}-\textbf{c.}, neither MelA$^+$ nor Sox9$^+$ cells are affected by the treatment, i.e., we mainly predict weights around 1. \textbf{d.} After 24h, we observe low weights in regions of high predicted apoptosis marker intensities (ClCasp3), especially at $t=24h$, where the observed cell counts suggest death predominantly in the MelA$^+$ cell cluster.}
    \label{fig:umap_ixazomib}
\end{figure}

\begin{figure}
    \centering
    \includegraphics[width=\textwidth]{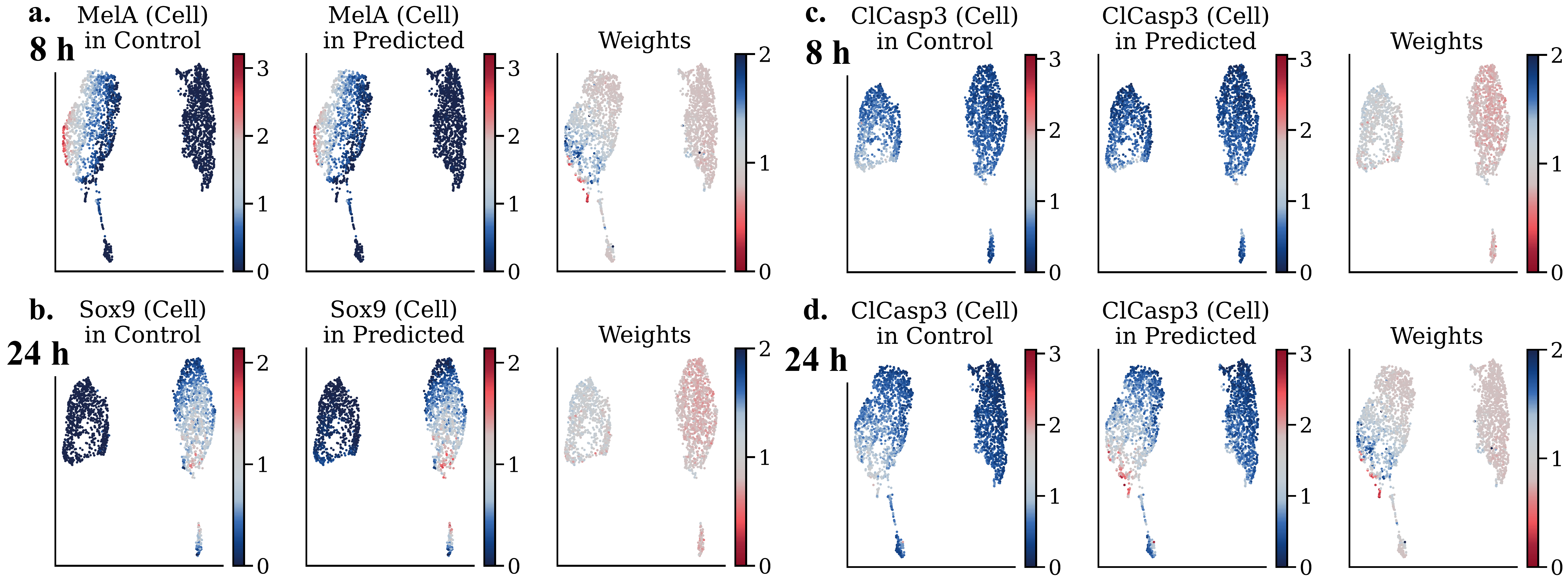}
    \caption{UMAP projections computed on control cells for Vindesine for $t=8h$, and $t=24h$, colored by protein marker intensities \textbf{a.} MelA and \textbf{b.} Sox9, markers for the two subpopulations, as well as ClCasp3, a marker for cell death, at \textbf{c.} 8h and \textbf{d.} 24h. The predicted weights (left) at \textbf{c.} 8h and \textbf{d.} 24h match the observed effects on each subpopulation, as initially only Sox9$^+$ cells are affected by treatment with Vindesine, and only after 24h MelA$^+$ cells show increased cell death. }
    \label{fig:umap_vindesine}
\end{figure}

As we lack ground truth for the correspondence of control and perturbed cells, we assess the biological meaningfulness of our predictions by comparing the weights to ClCasp3 and Ki67 intensity, the apoptosis and proliferation markers, respectively. Figures \ref{fig:umap_trametinib}, \ref{fig:umap_ixazomib} and \ref{fig:umap_vindesine} show UMAP projections computed on control cells for the drugs Trametinib, Ixazomib and Vindesine. 
In Figure~\ref{fig:umap_ixazomib}~c.,~d., and Figure~\ref{fig:umap_vindesine}~c.,~d., regions of low predicted weights accurately correspond to regions of increased ClCasp3 intensity. Additionally, we compare predicted weights between the two cell types, and contrast them with observed cell counts.

\section{Datasets} \label{sec:datasets}

We evaluate \textsc{NubOT} on several tasks including synthetic data as well as perturbation responses of single cells. In both settings, we are provided with unpaired measures $\mu$ and $\nu$ and aim to recover map $T$ which describes how source $\mu$ transforms into target $\nu$. While in the synthetic data setting we are provided with a ground truth matching, this is not the case for the single-cell data as measuring a cell requires destroying it. In the following, we describe generation and characteristics of both datasets, as well as introduce additional biological insights allowing us to shed light on the learned matching $T$.

\subsection{Synthetic Data} \label{sec:synth_datasets}

To evaluate \textsc{NubOT} in a simple and low-dimensional setup with known ground-truth, we generate synthetic example: We model a source population with clear subpopulation structure through a mixture of Gaussians. Next, we generate a second (target) population aligned to the source population. We then simulate an intervention to which the subpopulations respond differently, including different levels of growth and death.
Specifically, we generate batches of 400 samples with three clusters with different proportions before and after the intervention. Table~\ref{tab:toy_setup} shows the proportions of the three clusters in the source and target distribution, as well as the required weight-factor and the obtained results from \textsc{NubOT} and \textsc{ubOT GAN}.

\subsection{Single-Cell Data} \label{sec:cell_datasets}

\paragraph{Biological experiment.}
The single-cell dataset used in this work was generated by the a multiplexed microscopy technology called Iterative Indirect Immunofluorescence Imaging (4i) \citep{gut2018multiplexed}, which is capable of measuring the abundance and localization of many proteins in cells. By iteratively adding, imaging and removing fluorescently tagged antibodies, a multitude of protein markers is captured for each cell. Additionally, cellular and morphological characteristics are extracted from microscopical images, such as the cell and nucleus area and circularity.
This spatially resolved phenotypic dataset is rich in molecular information and provides insights into heterogeneous responses of thousands of cells to various drugs.
Measuring different morphological and signaling features captures pre-existing cell-to-cell variability which might influence perturbation effect, resulting in various different responses. Some of these markers are of particular importance, as they provide insights into the level of a cell's growth or death as well as subpopulation identity.
We utilized a mixture of two melanoma tumor cell lines (M130219 and M130429) at a ratio of 1:1. The cell lines can be differentiated by the mutually exclusive expression of marker proteins. The former is positive for Sox9, the latter for a set of four proteins which are all recognized by and antibody called MelA \citep{raaijmakers2015new}. Cells were seeded in a 384-well plate and incubated at 37C and 5\% CO2 overnight.  Next, the cells were exposed to multiple cancer drugs and Dimethyl sulfoxide (DMSO) as a vehicle control for 8h and 24h after which the cells were fixed and six cycles of 4i were performed TissueMAPS and the scikit-image library \citep{van2014scikit} were used to process and analyze the acquired images, perform feature extraction and quality control steps using semi-supervised random forest classifiers.

\paragraph{Data generation and processing.}
Our datasets contain high-dimensional single-cell data of control and drug-treated cells measured at two time points (8 and 24 hours). For both the 8h-dataset and the 24h-dataset, we normalized the extracted intensity and morphological features by dividing each feature by its 75th percentile, computed on the control cells. Additionally, values were transformed by a $log1p$ function ($x \leftarrow log(x+1)$). In total, our datasets consist of 48 features, of which 26 are protein marker intensities and the remaining 22 are morphological features. For each treatment, we have measured between 2000 and 3000 cells. For training the models, we perform a 80/20 train/test split. We trained all models on control and treated cells for each time step and each drug separately. The considered drugs as well as their inhibition type can be found in Table~\ref{tab:drugs}.

\begin{figure}
    \centering
    \includegraphics[width=1.05\textwidth]{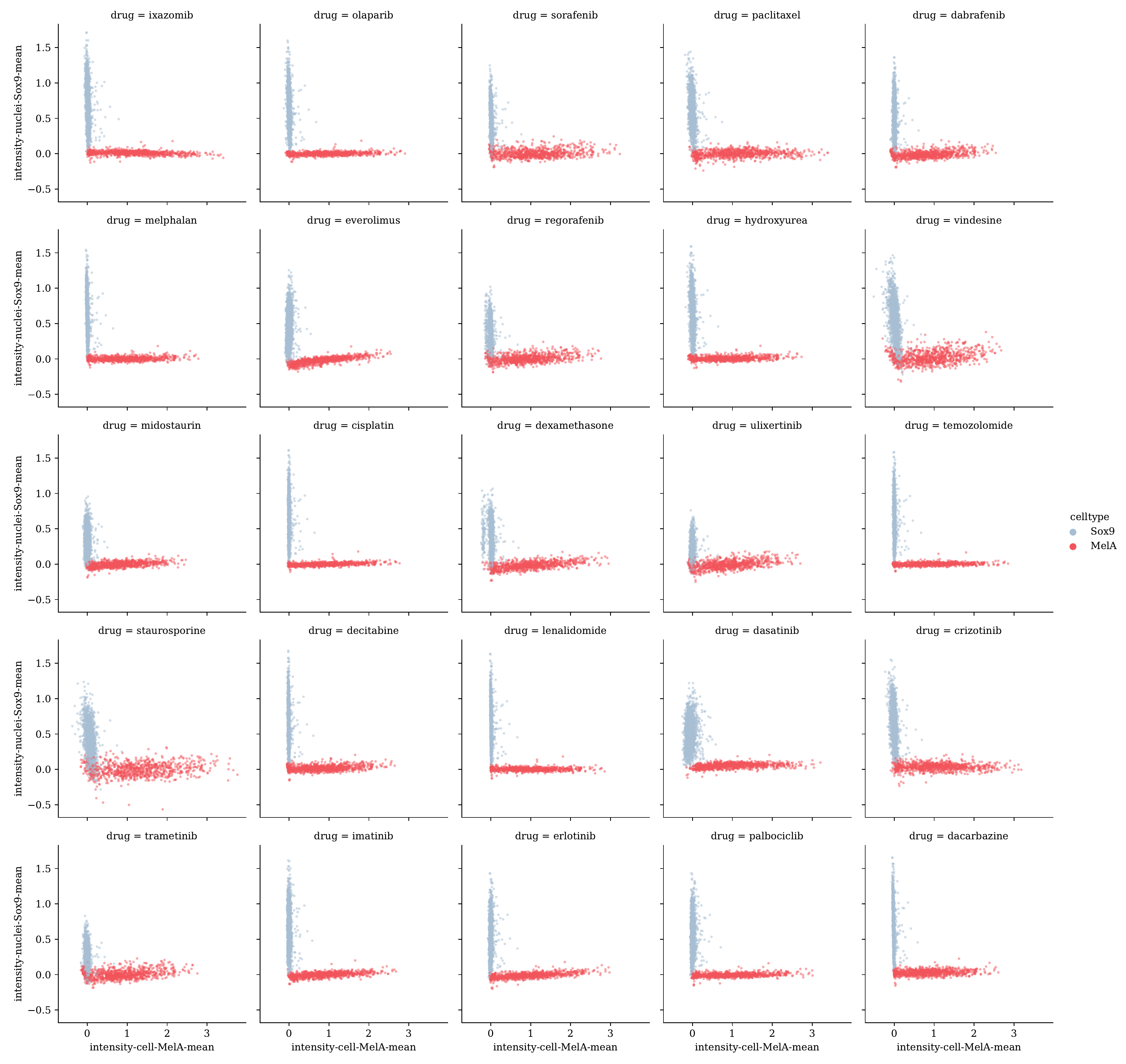}
     \caption{Classification of cells into cell types (MelA$^+$, Sox9$^+$) based on protein marker intensities of MelA and Sox9, for all drugs, at $t=8h$ \S~\ref{sec:cell_datasets}. Each tile represents one drug. MelA$^+$ cells colored in red, Sox9$^+$ in blue.}
    \label{fig:cell_types_check_8h}
\end{figure}

\begin{figure}
    \centering
    \includegraphics[width=\textwidth]{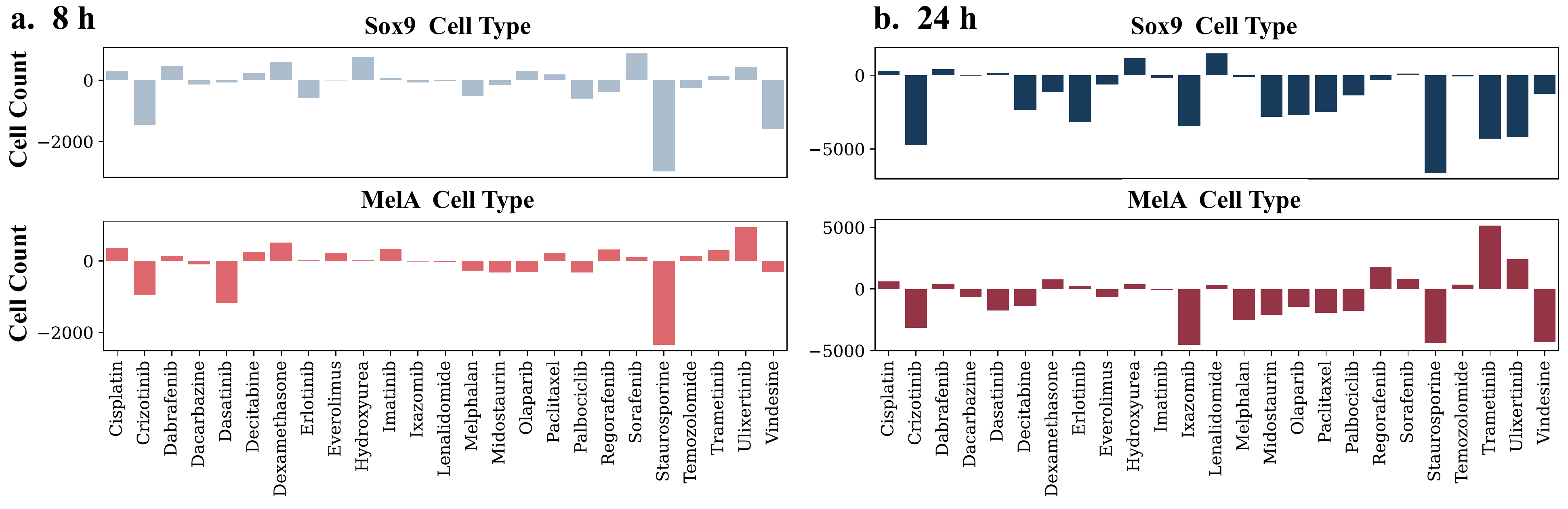}
    \caption{\looseness -1 Drug treatment-induced change in cell counts in the two cell types compared to the cell count of the respective cell types in the control condition. \textbf{a.} Cell count change for cell types Sox9$^+$ (top) and MelA$^+$ (bottom) at $t=8h$. \textbf{b.} Cell count change for cell types Sox9$^+$ (top) and MelA$^+$ (bottom) at $t=24h$.} 
    \label{fig:cell_counts_relative}
\end{figure}

\begin{figure}
    \centering
    \includegraphics[width=1.05\textwidth]{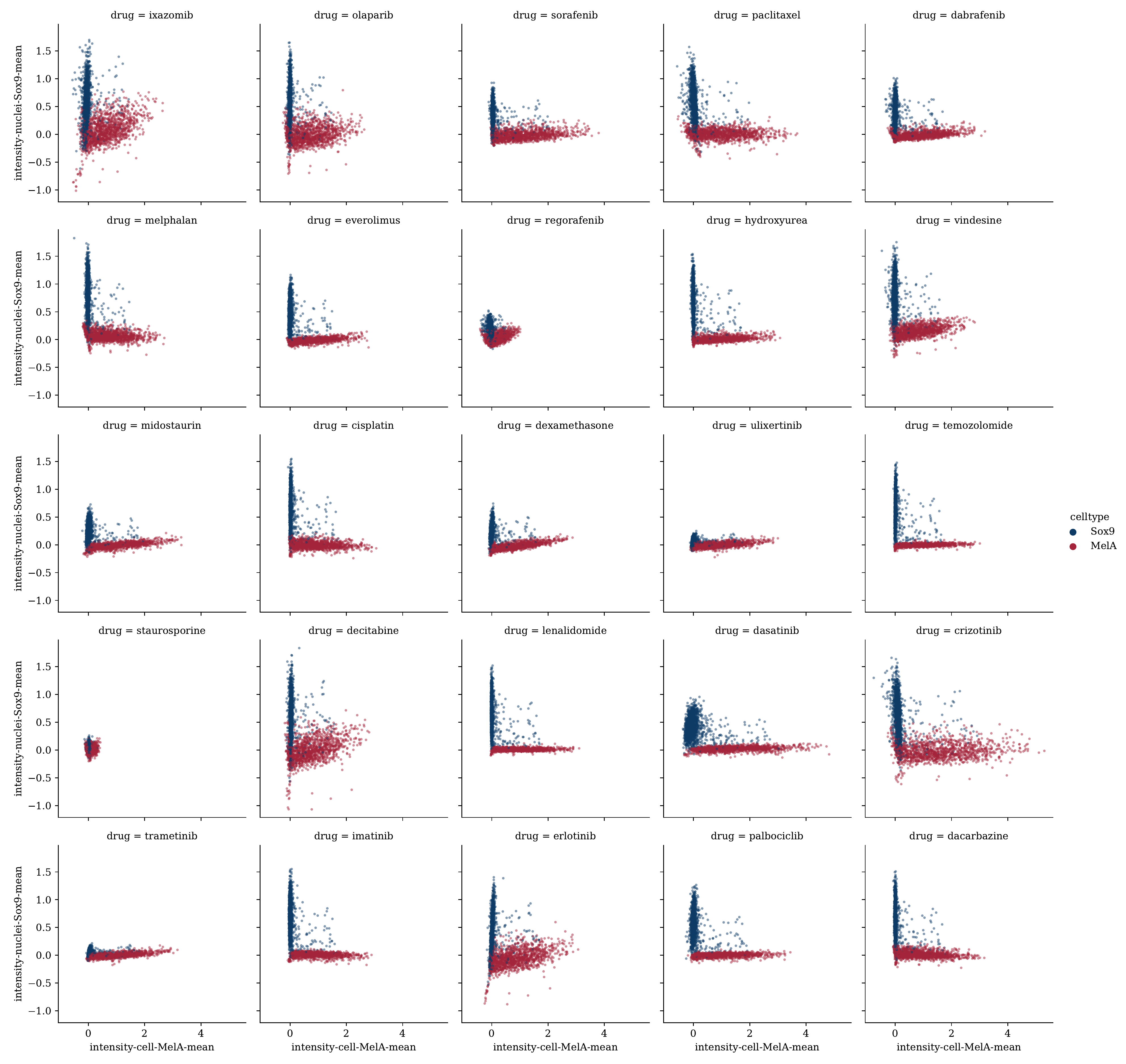}
    \caption{Classification of cells into cell types (MelA$^+$, Sox9$^+$) based on protein marker intensities of MelA and SOX9, for all drugs, at $t=24h$ \S~\ref{sec:cell_datasets}. MelA$^+$ cells colored in red, Sox9$^+$ in blue.}
    \label{fig:cell_types_check_24h}
\end{figure}

\begin{figure}
    \centering
    \includegraphics[width=\textwidth]{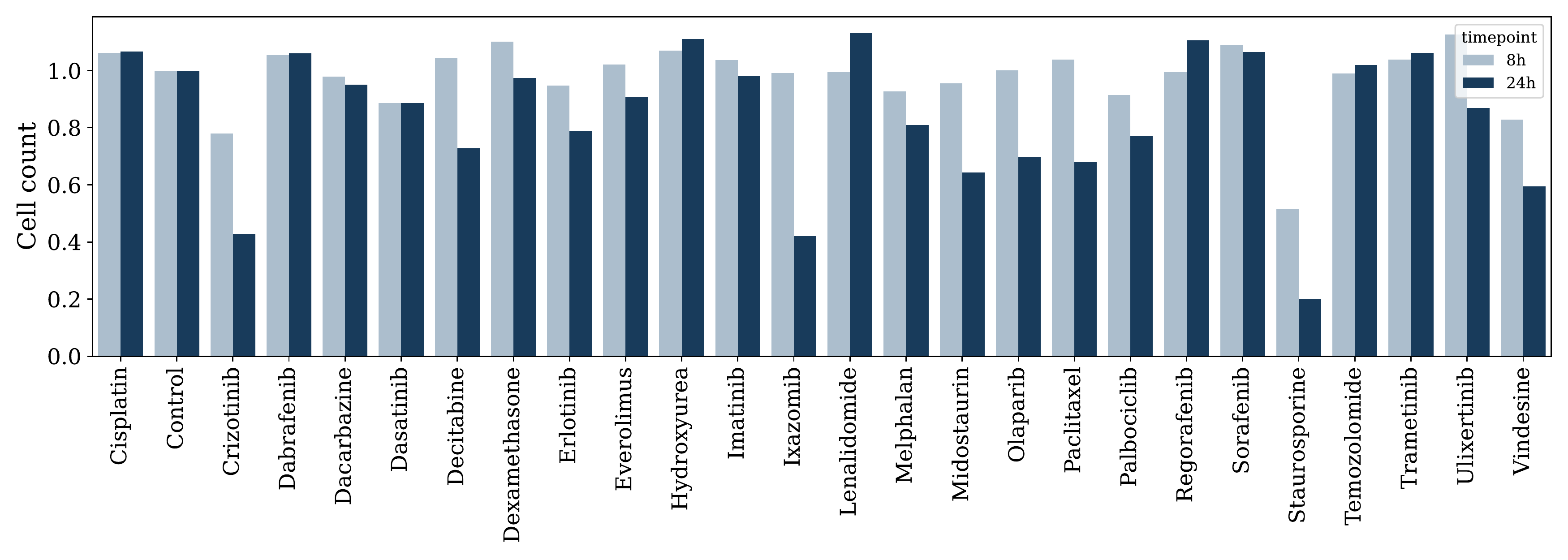}
    \caption{Observed cell counts of drug-treated cells normalized to control cell counts, per drug and time point. 8h treatment in light blue, 24h treatment in dark blue.}
    \label{fig:cell_counts_total}
\end{figure}

\begin{table}[]
    \centering
    \footnotesize
    \caption{\textbf{Overview of all treatments and their inhibition type considered in this work.} Abbreviations PROTi (Proteasome inhibitor), DNASynthi (DNA synthesis inhibitor), panKi (pan kinase inhibitor), ImmuneMod. (Immune modulatory compound), MTDisruptor (Microtubule disruptor), ApopInducer (Apoptosis inducer).}
    \label{tab:drugs}
    \begin{tabular}{lc||lc}
    \toprule
    \textbf{Drug Name} &            \textbf{Inhibitor Type} & \textbf{Drug Name} &     \textbf{Inhibitor Type} \\
    \midrule
    Ixazomib      &             PROTi &     Olaparib      &             PARPi \\
    Sorafenib     &              RAFi &     Paclitaxel    &       MTDisruptor \\
    Dabrafenib    &             BRAFi &     Melphalan     &         Alkylator \\
    Everolimus    &             mTORi &     Regorafenib   &             panKi \\
    Hydroxyurea   &          DNASynti &     Vindesine     &       MTDisruptor \\
    Midostaurin   &             panKi &     Cisplatin     &        Alkalyting \\
    Dexamethasone &        ImmuneMod. &     Ulixertinib   &              ERKi \\
    Temozolomide  &         Alkylator &     Staurosporine &       ApopInducer \\
    Decitabine    &           DNAMeti &     Lenalidomide  &        ImmuneMod. \\
    Dasatinib     &         SRCi-ABLi &     Crizotinib    &              METi \\
    Trametinib    &              MEKi &     Imatinib      &  KITi-PDGFRi-ABLi \\
    Erlotinib     &             EGFRi &     Palbociclib   &           CDK4/6i \\
    Dacarbazine   &         Alkylator \\
    \bottomrule
    \end{tabular}
\end{table}

\paragraph{Cell type assignment.}
We assigned M130219 and M130429 cells to the Sox9 and MelA cell types, respectively, by first training a two component Gaussian mixture model on the features 'intensity-cell-MelA-mean' and 'intensity-nuclei-Sox9-mean' of the control cells. Next, we used the aforementioned features and the labels provided by the mixture model to train a nearest neighbor classifier, which we then used to predict the cell type labels of the drug treated cells. The procedure was performed separately for the 8h- and 24h dataset. Results of the classification can be found in Figure \ref{fig:cell_types_check_8h} and Figure \ref{fig:cell_types_check_24h} respectively.

\section{Experimental Details} \label{sec:exp_details}

\textsc{NubOT} consists of several modules and its performance is compared against several baselines. In the following, we provide additional background on experimental details, including a description of the evaluation metrics and baselines considered, as well as further information on the parameterization and hyperparameter choices made for \textsc{NubOT}.

\subsection{Evaluation Metrics} \label{sec:eval_metrics}

We evaluate our model by analyzing the distributional similarity between the predicted and observed perturbed distribution. For this, we compute the kernel maximum mean discrepancy (MMD) \citep{gretton2012kernel}. To take the mass variation into consideration, we compute a weighted version of MMD, by weighting each predicted point by its associated normalized weight.

\subsection{Baselines} \label{sec:baselines}

We compare \textsc{NubOT} against several baselines, comprising a balanced OT-based method \citep[\textsc{CellOT}]{bunne2021learning} and an unbalanced OT-based method \citep[\textsc{NubOT}]{yang2018scalable}, i.e., current state-of-the-art methods as well as ablations of our work. In the following, we briefly motivate and introduce each baseline.

\paragraph{\textsc{CellOT}.}
By introducing reweighting functions $\eta$ and $\zeta$, \textsc{NubOT} recovers a balanced problem parameterized by dual potentials $f$ and $g$. An important ablation study to consider is thus to compare its performance to its balanced counterpart. Ignoring the fact that the original problem includes cell death and growth, and thus varying cell numbers, we apply ideas developed in \citet{makkuva2020optimal, bunne2021learning} and learn a balanced OT problem via duals $f$ and $g$. These duals are parameterized by two ICNNs and optimized in objective \eqref{eq:not-minmax} via an alternating min-max scheme. 

\paragraph{\textsc{ubOT GAN}.}
Using \eqref{eq:ub-ot}, \citet{yang2018scalable} propose to model mass variation in unbalanced OT via a relaxation of the marginals.
Similar to \citet{fan2021scalable}, \citet{yang2018scalable} reformulate the constrained Monge problem~\eqref{eq:monge} as a saddle point problem with Lagrange multiplier $h$ for the constraint $T_\sharp \mu = \nu$, i.e.,
\begin{align*}
    \sup_h \inf_T &\int_{\gX} c(x, T(x)) \mu(x) d x+\int_{\gX} h(y)\left(\nu-T_{\sharp} \mu\right) d y \\
    = &\int_{\gX}[c(x, T(x))-h(T(x))] \mu(x) d x+\int_{\gX} h(y) \nu(y) d y,
\end{align*}
parameterizing $T$ and $h$ via neural networks. 
To allow mass to be created and destroyed, \citet{yang2018scalable} introduce scaling factor $\xi: \mathcal{X} \rightarrow \mathbb{R}^+$, allowing to scale mass of each source point $x_i$. The optimal solution then needs to balance the cost of mass and the cost of transport, potentially measured through different cost functions $c_1: \mathcal{X} \times \mathcal{Y} \rightarrow \mathbb{R}^+$ (cost of mass transport) and $c_2: \mathbb{R}^+ \rightarrow \mathbb{R}^+$ (cost of mass variation). Parameterizing the transport map $T_{\theta}$, the scaling factors $\xi_{\phi}$, and the penalty $h_{\omega}$ with neural networks, the resulting objective is
\begin{equation*} \label{eq:gan-loss}
    l(\theta, \phi, \omega) \coloneqq \frac{1}{n} \sum_{i=0}^{n} \left[ c_1(x_i,T_{\theta}(x_i)) \xi_{\phi}(x_i) + c_2(\xi_{\phi}(x_i)) + \xi_{\phi}(x_i) h_{\omega}(T_{\theta}(x_i) - \Psi^*(h_{\omega}(y_i) \right],
\end{equation*}
with $\Psi^*$ approximating the divergence term of the relaxed marginal constraints (see \eqref{eq:ub-ot}), and is optimized via alternating gradient updates. 

\paragraph{\textsc{Identity}.}
A trivial baseline is to compare the predictions to a map which does not model any perturbation effect. The \textsc{Identity} baseline thus models an identity map and provides an \emph{upper bound} on the overall performance, also considered in \citet{bunne2021learning}.

\paragraph{\textsc{Observed}.}
In a similar fashion we might ask for a \emph{lower bound} on the performance. As a ground truth matching is not available, we can construct a baseline for a comparison on a distributional level by  comprising a different set of observed perturbed cells, which only vary from the true predictions up to experimental noise. The closer a method can approach the \textsc{Observed} baseline, the more accurate it fits the perturbed cell population.

\subsection{Hyperparameters} \label{sec:hyperparams}
We parameterize the duals $f$ and $g$ using ICNNs with 4 hidden layers, each of size 64, using ReLU as activation function between the layers. We choose the identity initialization scheme introduced by \citet{bunne2022supervised} such that $\nabla g$ and $\nabla f$ resemble the identity function in the first training iteration.
As suggested by \citet{makkuva2020optimal}, we relax the convexity constraint on ICNN $g$ and instead penalize its negative weights $W^z_l$
\begin{equation*}
    R\left(\theta\right)=\lambda \sum_{W^z_l \in \theta}\left\|\max \left(-W^z_l, 0\right)\right\|_{F}^{2}.
\end{equation*}
The convexity constraint on ICNN $f$ is enforced after each update by setting the negative weights of all $W^z_l \in \theta_f $ to zero. Duals $g$ and $f$ are trained with an alternating min-max scheme where each model is trained at the same frequency.
Further, both reweighting functions $\eta$ and $\zeta$ are represented by a multi-layer perceptron (MLP) with two hidden layers of size 64 for the single-cell and of size 32 for the synthetic dataset, with ReLU activation functions. The final output is further passed through a softplus activation function as we do not assume negative weights.
For the unbalanced Sinkhorn algorithm, we choose an entropy regularization of $\varepsilon = 0.005$ and a marginal relaxation penalty of $0.05$.
We use both Adam for pairs $g$ and $f$ as well as $\eta$ and $\zeta$ with learning rate $10^{-4}$ and $10^{-3}$ as well as $\beta_1 = 0.5$ and $\beta_2 = 0.9$, respectively. We parameterize both baselines with networks of similar size and follow the implementation proposed by \citet{yang2018scalable} and \citet{bunne2021learning}.

\section{Reproducibility}
The code will be made public upon publication of this work.

\end{document}